\documentclass[twoside]{article}
\usepackage{amsfonts, amsmath, amsthm, amssymb}
\usepackage{natbib}
\usepackage[lined,linesnumbered,boxruled,noend]{algorithm2e}
\usepackage{graphicx}
\usepackage{hyperref}

\usepackage[accepted]{aistats2016}

\newtheorem{lemma}{Lemma}
\newtheorem*{lemma*}{Lemma}
\newtheorem{theorem}{Theorem}
\newtheorem{definition}{Definition}

\newtheorem{appcond}{Condition}
\newtheorem{applemma}{Lemma}
\newtheorem*{applemma*}{Lemma}
\newtheorem{apptheorem}{Theorem}

\newcommand{\porl}{EEPORL} %\algfont{EEPORL}}

\begin{document}
	
	
	
	\twocolumn[
	
	\aistatstitle{A PAC RL Algorithm for Episodic POMDPs}
	
	\aistatsauthor{ Zhaohan Daniel Guo \And Shayan Doroudi \And Emma Brunskill }
	
	\aistatsaddress{ Carnegie Mellon University \\ 5000 Forbes Ave \\ Pittsburgh PA 15213, USA \And Carnegie Mellon University \\ 5000 Forbes Ave \\ Pittsburgh PA 15213, USA \And Carnegie Mellon University \\ 5000 Forbes Ave \\ Pittsburgh PA 15213, USA} ]
	
	\begin{abstract}
		Many interesting real world domains involve reinforcement 
		learning (RL) in partially observable environments. Efficient 
		learning in such domains is important, but 
		existing sample complexity bounds for partially observable RL 
		are at least exponential in the
		episode length. 
		We give, to our knowledge, the first partially observable RL algorithm 
		with a polynomial bound on the number of episodes 
		on which the algorithm may not achieve near-optimal performance. 
		Our algorithm is suitable for an important class 
		of episodic POMDPs. Our approach builds on recent advances in method of moments for latent variable model estimation. 
	\end{abstract}
	
	\section{INTRODUCTION}
	
	A key challenge in artificial intelligence is how to effectively learn to make a sequence of good decisions in stochastic, 
	unknown environments. Reinforcement learning (RL) is a subfield 
	specifically focused on how agents can learn to make 
	good decisions given feedback in the form of a reward 
	signal. In many important applications such as robotics, education, and healthcare, the agent cannot directly observe the state of the environment responsible for generating the reward signal, and instead only receives incomplete or noisy observations.
	
	One important measure of an RL algorithm is its sample 
	efficiency: how much data/experience is needed to compute a 
	good policy and act well. One way to measure sample complexity is given by the Probably Approximately Correct framework; an RL algorithm is said to be PAC if with high probability, it selects a near-optimal action on all but a number of steps (the sample complexity) which is a polynomial function of the problem parameters. There 
	has been substantial progress on PAC RL for the fully observable 
	setting~\citep{brafman2003r, strehl2005theoretical, kakade2003sample, strehl2012incremental, lattimore2012pac}, but to our knowledge 
	there exists no published work on PAC RL algorithms for 
	partially observable settings. 
	
	This lack of work on PAC partially observable RL is perhaps 
	because of the additional challenge introduced by the partial observability of the environment. In fully observable settings, the world is often assumed to behave as a Markov decision process (MDP). An elegant approach for proving that a RL algorithm for MDPs is PAC is to compute finite sample error bounds on the MDP parameters. However, because the states of a partially observable MDP (POMDP) are hidden, the naive approach of directly 
	treating the POMDP as a history-based 
	MDP yields a state space that grows exponentially 
	with the horizon, rather 
	than polynomial in all POMDP parameters~\citep{even2005reinforcement}. 
	
	On the other hand, there has been substantial 
	recent interest and progress 
	on method of moments and spectral 
	approaches for modeling partially observable systems~\citep{anandkumar2012method,anandkumar2014tensor,HKZ09,littman2001predictive,boots2011closing}. 
	The majority of this work has focused on inference 
	and prediction, with little work tackling the 
	control setting. Method of moments 
	approaches to latent variable estimation are of particular 
	interest because for a number of models they obtain 
	global optima and 
	provide finite sample guarantees on the accuracy of the learned
	model parameters.
	
	Inspired by the this work, we propose a POMDP RL algorithm that is, to our knowledge, the first PAC POMDP RL algorithm for episodic domains (with no restriction on the policy class).  Our algorithm is 
	applicable to a restricted but important class of 
	POMDP settings, which include but are not limited to information gathering POMDP RL 
	domains such as preference elicitation~\citep{boutilier2002pomdp}, dialogue management 
	slot-filling domains~\citep{ko2010structured}, and medical diagnosis before decision 
	making~\citep{amato2012}. Our work builds 
	on method of moments inference techniques, but requires 
	several non-trivial extensions to tackle the control 
	setting. In particular, there is a subtle issue of latent 
	state alignment: if the models for each action are learned 
	as independent hidden Markov models (HMMs), then it 
	is unclear how to solve the correspondence issue across 
	latent states, which is essential for performing planning 
	and selecting actions. Our primary contribution is 
	to provide a theoretical analysis of our proposed algorithm, 
	and prove that it is possible to obtain near-optimal 
	performance on all but a number of episodes that scales as a 
	\textit{polynomial} function of the POMDP parameters.
	Similar to most fully observable PAC RL 
	algorithms, directly instantiating our bounds would yield 
	an impractical number of samples for a real application.  
	Nevertheless, we believe understanding the sample complexity 
	may help to guide the amount of data required for a task, 
	and also similar to PAC MDP RL work, may motivate new 
	practical algorithms that build on these ideas. 
	
	\section{BACKGROUND AND RELATED WORK}
	
	The inspiration for pursuing PAC bounds for POMDPs came about from the success of PAC bounds for MDPs ~\citep{brafman2003r, strehl2005theoretical, kakade2003sample, strehl2012incremental, lattimore2012pac}. While algorithms have been developed for POMDPs with finite sample bounds~\citep{peshkin2001bounds,even2005reinforcement}, unfortunately these bounds are not PAC as they have an exponential dependence on the horizon length.
	
	Alternatively, Bayesian methods~\citep{ross2011bayesian, doshi2012bayesian} are very popular for solving POMDPs. For MDPs, there exist Bayesian methods that have PAC bounds~\citep{kolter2009near,asmuth2009bayesian}; however there have been no PAC bounds for Bayesian methods for POMDPs. That said, Bayesian methods are optimal in the Bayesian sense of making the best decision given the posterior over all possible future observations, which does not translate to a frequentist finite sample bound.
	
	We build on method of moments (MoM) work for estimating HMMs~\citep{anandkumar2012method} in order to provide a finite sample bound for POMDPs. MoM is able to obtain a global optimum, and has finite sample bounds on the accuracy of their estimates, unlike the popular Expectation-Maximization (EM) that is only guaranteed to find a local optima, and offers no finite sample guarantees. MLE approaches for estimating HMMs~\citep{abe1992computational} also unfortunately do not provide accuracy guarantees on the estimated HMM parameters. As POMDP planning methods typically require us to have estimates of the underlying POMDP parameters, it would be difficult to use such MLE 
	methods for computing a POMDP policy and providing a finite sample guarantee\footnote{\citet{abe1992computational}'s 
		MLE approach guarantees that the estimated probability over $H$-length observation sequences 
		has a bounded KL-divergence from the true probability of the sequence under the true 
		parameters, which is expressed as a function of the number of underlying data samples 
		used to estimate the HMM parameters. We think it may be possible to use such 
		estimates in the control setting when modeling hidden state control systems as 
		PSRs, and employing a forward search approach to planning; 
		however, there remain a number of subtle issues to address to ensure such an 
		approach is viable and we leave this as an interesting direction for future work.}.
	
	Aside from the MoM method in \citet{anandkumar2012method}, another popular spectral method involves using Predictive State Representations (PSRs) \citep{littman2001predictive,boots2011closing}, to directly tackle the control setting; however it only has asymptotic convergence guarantees and no finite sample analysis. There is also another method of moments approach to transfer across a set of bandits tasks, but the latent variable estimation problem is substantially simplified because the state of the system is 
	unchanged by the selected actions~\citep{mab2013}.

	Fortunately, due 
	to the polynomial finite sample bounds from MoM, we 
	can achieve a PAC (polynomial) sample complexity bound for POMDPs.

\section{PROBLEM SETTING}
We consider a partially observable Markov decision process (POMDP) 
which is described as the tuple $(S,A,R,T,Z,b,H)$ 
where we have a set of discrete states $S$, discrete actions $A$, discrete observations $Z$, discrete rewards $R$, initial belief $b$ (more details below), and episode length $H$. The transition model is represented by a set of $|A|$ matrices 
 $T_a(i,j) : |S| \times |S|$ where the $(i,j)$-th entry is the probability 
of transitioning from $s_i$ to $s_j$ under action $a$. 
With a slight abuse of notation, we use $Z$ to denote both the 
finite set of observations and the observation model captured 
by the set of $|A|$ observation matrices, $Z_a$ where the $(i,j)$-th 
entry represents the probability of observing $z_i$ given the 
agent took action $a$ and transitioned to state $s_j$. We 
similarly do a slight abuse of notation and let $R$ denote 
both the finite set of rewards, and the reward matrices 
$R_a$ where the $(i,j)$-th entry in a matrix denotes the probability 
of obtaining reward $r_i$ when taking action $a$ in state $s_j$. Note that in our setting we also treat the reward as an additional observation\footnote{In 
planning problems the reward is typically a real-valued 
scalar, but in PORL we must learn the reward model. This 
requires assuming some mapping between states and rewards. 
For simplicity we assume multinomial distribution over a discrete set of 
rewards. Note that we can always 
discretized a real-valued reward into a finite set of 
values with bounded error on the resulting value function 
estimates, and our choice makes very little restrictions 
on the underlying setting.}.

The objective in POMDP planning is to compute a policy $\pi$ that 
achieves a large expected sum of future rewards, where $\pi$ is a mapping from histories of prior sequences of actions, observations, and rewards, to actions. In many 
cases we capture prior histories using a sufficient statistic called the belief $b$ where 
$b(s)$ represents the probability of being in a particular 
state $s$ given the prior history of actions, observations and rewards. 
One popular method for POMDP planning involves representing the 
value function by a finite set of $\alpha$-vectors, where 
$\alpha(s)$ represents the expected sum of future rewards of 
following the policy associated with the $\alpha$-vector from 
initial state $s$. POMDP planning then proceeds 
by taking the first action associated with the policy of the $\alpha$-vector which 
yields the maximum expected value for the current belief state, 
which can be computed for a particular $\alpha$-vector using 
the dot product $\langle b, \alpha \rangle$. 

In the 
reinforcement learning setting, the transition, observation, 
and/or reward model parameters are initially unknown. 
The goal is to learn a policy that achieves large sum of rewards in the environment without advance knowledge of 
how the world works. 

We make the following assumptions about the domain and 
problem setting:
\begin{enumerate}
\item We consider episodic, finite horizon partially observable 
RL (PORL) settings
\item It is possible to achieve a non-zero probability of 
being in any state in two steps from the initial belief. 
\item For each action $a$, the transition matrix $T_a$ is full rank, 
and the observation matrix $Z_a$ and reward matrix $R_a$ are 
full column rank.
\end{enumerate}
The first assumption on the setting is satisfied by many real world situations involving an agent 
repeatedly doing a task: for example, an agent may sequentially interact with many different customers each for a finite amount 
of time. 
The key restrictions on the setting are captured in 
assumptions 2 and 3. Assumption 2 is similar to a mixing assumption and is necessary in order for MoM to estimate dynamics for all states. Assumption 3 is necessary for MoM to uniquely determine the transition, observation, and reward dynamics. The second assumption may sound 
quite strong, as in some POMDP settings states are 
only reachable by a complex sequence of carefully chosen 
actions, such as in robotic navigation or video games. % may switch examples as can treat video as MDP
However, assumption 2 is commonly satisfied in many important POMDP settings that primarily involve information gathering. 
For example, in preference elicitation or user modeling, 
POMDPs are commonly used to identify the, typically static,  
hidden intent or preference or state of the user, before 
taking some action based on the resulting information~\citep{boutilier2002pomdp}. Examples of this include dialog systems~\citep{ko2010structured}, medical diagnosis and 
decision support~\citep{amato2012}, and even human-robot collaboration preference 
modeling~\citep{nikolaidis2015efficient}.  In such settings, the belief commonly starts out non-zero over all possible user states, and 
slowly gets narrowed down over time. The third assumption is 
also significant, but is still satisfied by an important class of 
problems that overlap with the settings captured by 
assumption 2. Information gathering POMDPs where the state is 
hidden but static automatically satisfy the full rank assumption 
on the transition model, since it is an identity matrix. 
Assumption 3 on the observation and reward matrices imply 
that the cardinality of the set of observations (and rewards) 
is at least as large as the size of the state space. 
A similar assumption has been made in many latent variable 
estimation settings (e.g.~\citep{anandkumar2012method,anandkumar2014tensor,Song:2010fk}) 
including in the control setting~\citep{boots2011closing}. Indeed, when the observations 
consist of videos, images or audio signals, this assumption is typically satisfied~\citep{boots2011closing}, and such signals are very 
common in dialog systems and the user intent and modeling 
situations covered by assumption 2.  Satisfying that the 
reward matrix has full rank is typically trivial as the 
reward signal is often obtained by discretizing a real-valued 
reward. Therefore, while 
we readily acknowledge that our setting does not 
cover all generic POMDP reinforcement learning settings, we 
believe it does cover an important class of problems 
that are relevant to real applications.

% EB: still might sound weird for the first two states. 
% what if can instantly diagnose

% EB: discuss why not infinite horizon. do we use a discount?
%     could we use our approach with the homing procedure 
%     to address infinite horizon? seems hard, don't know 
%     initial belief covers all states

% EB: our results apply to discrete values rewards
%     treast as observations. Interesting extension to 
%     cover real valued rewards but differ to future 

% could there be methods that do direct value estimates 
% with bounds for POMDPs?

\begin{algorithm}[h!]
\SetKwInOut{Input}{input}
\Input{$S,A,Z,R,H,N,c,\pi_{rest}$}
Let $\pi_{explore}$ be the policy where $a_1,a_2$ are uniformly random, and $p(a_{t+2}|a_t) = \frac{1}{1 + c|A|}(\mathbf{I} + c\mathbf{1}_{|A|\times|A|})$ \;
$X \gets \emptyset$ \;
\tcp{Phase 1:}
\For{episode $i \gets 1$ \KwTo $N$}{
Follow $\pi_{explore}$ for 4 steps \;
Let $x_t = (a_t, r_t,z_t,a_{t+1})$ \;
$X \gets X \cup \{(x_1, x_2, x_3)\}$ \;
Execute $\pi_{rest}$ for the rest of the steps \;
}
\tcp{\textbf{Phase 2:}}
Get $\widehat{T},\widehat{O},\widehat{w}$ for the induced $HMM$ from $X$ through our extended MoM method \;
Using the labeling from Algorithm \ref{alg:labelactions} with $\widehat{O}$, compute estimated POMDP parameters.\;
Call Algorithm \ref{alg:findpolicy} with estimated POMDP parameters to estimate a near optimal policy $\widehat{\pi}$ \;
Execute $\widehat{\pi}$ for the rest of the episodes \;
\caption{\porl}
\label{alg:pac}
\end{algorithm}

\begin{algorithm}[h!]
\SetKwInOut{Input}{input}
\Input{$\widehat{O}$}
\ForEach{column $i$ of $\widehat{O}$}{
Find a row $j$ such that $\widehat{O}(i,j) \geq 
\frac{2}{3|R||Z|}$ \;
Let the observation associated with row $j$ be $(a,r',z',a')$, label column $i$ with $(a,a')$ \;
}
\caption{LabelActions}
\label{alg:labelactions}
\end{algorithm}

\begin{algorithm}[h!]
\SetKwInOut{Input}{input}
\Input{$\widehat{b}(s_{(a_0,a_1)})$, $\widehat{p}(z | a, s_{(a,a')})$, $\widehat{p}(r | s_{(a,a')},a')$, $\widehat{p}(s_{(a',a'')} | s_{(a,a')}, a')$
}
$\forall a_-,a \in A, \quad \Gamma^{a_-,a}_1 = \{ \widehat{\beta}_1^a(s_{(a_-,a)}) \}$ \;
\For{$t \gets 2$ \KwTo $H$}{
$\forall a,a' \in A, \Gamma^{a,a'}_t = \emptyset$ \;
\For{$a,a' \in A$}
{
\For{$f_t(r,z) \in (|R|\times|Z| \rightarrow \Gamma_{t-1}^{a,a'})$}{
\tcp{all mappings from an observation pair to a previous $\beta$-vector}
$\forall a_- \in A, \Gamma^{a_-,a}_t = \Gamma^{a_-,a}_t \cup \{ \beta_t^{a,f_t}(s_{(a_-,a)}) \}$ \;
}
}
}
Return $\arg\max_{a_0,a_1,\beta_H(s_{(a_0,a_1)})) \in \Gamma_H^{a_0,a_1}} (\widehat{b} \cdot \beta_H)$  \;
\caption{FindPolicy}
\label{alg:findpolicy}
\end{algorithm}

\begin{figure}[h]
	\centering
	%\vspace{.3in}
	\includegraphics[width=0.46\textwidth,natheight=160,natwidth=437]{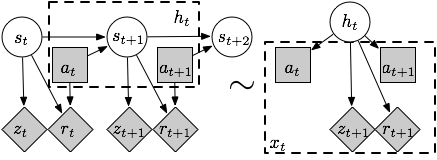}
	%\vspace{.3in}
	\caption{POMDP (left) analogous to induced HMM (right). Gray nodes show fully observed variables, whereas white nodes show latent states.}
	\label{fig:iinduce}
	\vspace{-1em}
\end{figure}

\section{ALGORITHM}
Our goal is to create an algorithm that can achieve 
near optimal performance from the initial belief on each episode. 
Prior work has shown that the error in the POMDP value 
function is bounded when using model parameter 
estimates that themselves have bounded error~\citep{ross2009sensitivity,fard2008variance}; 
however, this work takes a 
sensitivity analysis 
perspective, and does not address how such model estimation 
errors themselves could be computed or bounded.\footnote{
~\citet{fard2008variance} assume 
that labels of the hidden states are provided, which 
removes the need for latent variable esimtation.} 

In contrast, many PAC RL algorithms for MDPs have shown that exploration is critical in order to get enough data to estimate the model parameters. However in MDPs, algorithms can directly observe how many times every action has been tried in every state, and can use this information to steer exploration towards less explored areas. In partially observable settings it is more challenging, as the state itself is hidden, and so it is not possible to directly observe the number of times an action has been tried in a latent state. Fortunately, recent advances in method of moments (MoM) estimation procedures for latent variable estimation (see e.g.~\citep{anandkumar2012method,anandkumar2014tensor}) have demonstrated that in certain uncontrolled settings, including many types of hidden Markov models (HMMs), it is still possible to achieve accuracy estimates of the underlying latent variable model parameters as a function of the amount of data samples used to perform the estimation. For some intuition about this, consider starting in a belief state $b$ which has non-zero probability over all possible states. If one can repeatedly take the same action $a$ from same belief $b$, given a sufficient number of samples, we will have actually taken action $a$ in each state many times (even if we don't know the specific instances on which action $a$ was taken in a state $s$).

The control setting is more subtle than the uncontrolled 
setting which has been the focus of the majority of recent MoM spectral learning research, because we wish to estimate not just the transition 
and observation models of a HMM, but to estimate the 
POMDP model parameters. Our ultimate interest 
is in being able to select good actions. A naive approach 
is to independently learn the transition, observation, and reward parameters for each separate action, by restricting the POMDP to only execute a single action,
 thereby turning the POMDP  into an HMM. 
However, this simple aproach fails because 
the returned parameters can correspond to a different labeling of the hidden states. 
For example, 
the first column of the transition matrix for action $a_1$ may actually correspond to the state $s_2$,
 while the first column of the transition matrix for action $a_2$ may 
truly correspond to $s_5$. We require that the labeling must be consistent for all actions since we wish to compute what happens when different actions are executed consecutively. An unsatisfactory way to match up the labels for different actions is by requiring that the initial belief state have probabilities that are unique and well separated per state. Then we can use the estimated initial belief from each action to match up the labels.
%~\footnote{If one knows that $b(s)$ values are unique and well separated for each state $s$, then if these are estimated to a sufficient degree of precision, it will be possible to do the alignment between two believe vectors $b_1$ and $b_2$ which are known to be permutations of the same underlying belief state vector.}. 
However, this is a very strong assumption on the starting belief 
state which is unlikely to be realized. 

To address this challenge of mismatched labels, we transform our 
POMDP into an induced 
HMM (see Figure~\ref{fig:iinduce}) by fixing the policy to $\pi_{explore}$ (for a few steps, during a certain 
number of episodes), and create an alternate hidden state representation 
that directly solves the problem of alignment of hidden states across actions. 
Specifically, we make the hidden state at time $t$ of the induced HMM, denoted by 
$h_t$, equal to the tuple of the action at time step $t$, the next state, and 
the subsequent action, $h_t = (a_t,s_{t+1},a_{t+1})$. We denote the 
observations of the induced HMM by $x$, and the observation associated 
with a hidden state $h_t$ is the tuple $x_t = (a_t,r_t,z_t,a_{t+1})$. 
Figure~\ref{fig:iinduce} shows how the graphical model of our 
original POMDP is related to the graphical model of the induced HMM. 
In making this transformation, our resulting HMM still satisfies the 
Markov assumption: the next state is only a function of the prior state,  and the observation is only a function of the current state. But, 
this transformation also has the desired property that it is now 
possible to directly align the identity of states across selected 
actions. This is because HMM parameters now depend on both state and action, so there is a built-in correlation between different actions. We will discuss 
this more in the theoretical analysis. % EB: could put more here?

We are now ready to  describe our algorithm for episodic finite horizon reinforcement learning in 
POMDPs, \porl\ (Explore then Exploit Partially Observable RL, which is shown in Algorithm~\ref{alg:pac}). 
Our algorithm is model-based and proceeds in two phases. In the first phase, it performs exploration to collect samples of trying different actions in different (latent) states. After the first phase completes, we extend a MoM approach~\citep{anandkumar2012method} to compute estimates of the induced HMM parameters. We use these estimates to obtain a near-optimal policy. 

\subsection{Phase 1}

The first phase consists of the first $N$ episodes. 
Let $\pi_{explore}$ be a fixed open-loop policy for the 
first four actions of an episode. In $\pi_{explore}$ 
 actions $a_1,a_2$ are selected uniformly at random, 
and $p(a_{t+2}|a_t) = \frac{1}{1 + c|A|}(\mathbf{I} + c\mathbf{1}_{|A|\times|A|})$ \; 
where $c$ can be any positive real number. For our proof, we pick $c = O(1/|A|)$. 
Note that $\pi_{explore}$  only depends on previous actions and not on any observations. The definition of $p(a_{t+2}|a_t)$ for what will work for the proof only requires it to be full-rank and having some minimum probability over all actions. We chose a perturbed identity matrix for simplicity. Since $\pi_{explore}$ is a fixed policy, the POMDP process reduces to a HMM for these first four steps. During these steps we store the observed experience as 
$(x_1, x_2, x_3)$, where $x_t = (a_t,r_t,z_t,a_{t+1})$ is an observation of our previously defined induced HMM. 
The algorithm then follows policy $\pi_{rest}$ for the remaining 
steps of the episode. All of these episodes will be considered 
as potentially non-optimal, and so the choice of $\pi_{rest}$ 
does not impact the theoretical analysis. However, empirically 
$\pi_{rest}$ could be constructed to encourage near optimal 
behavior given the observed data collected up to the current 
episode. 

\subsection{Parameter Estimation}

After Phase 1 completes, we have $N$ samples of the tuple $(x_1, x_2, x_3)$. We then apply our extension to the MoM algorithm for HMM parameter estimation by \citet{anandkumar2012method}. Our extension computes estimates and bounds on the transition model $\widehat{T}$ which is not computed in the original method. To summarize, this 
procedure yields 
%Our extension involves computing $\widehat{T}$ (the original method only had $
% This returns 
an estimated transition matrix $\widehat{T}$, observation matrix $\widehat{O}$, and belief vector $\widehat{w}$ for the induced HMM. The belief $\widehat{w}$ is over the second hidden state, $h_2$.

As mentioned before as one major challenge, labeling of the states $h$ 
of the induced HMM is arbitrary; however it is consistent between 
$\widehat{T},\widehat{O},\widehat{w}$ since this is a single HMM inference problem. Recall that a hidden state in our induced HMM is defined as $h_t=(a_t,s_{t+1},a_{t+1})$. Since the actions are fully observable, it is possible to label each state $h=(a,s',a')$  (i.e. the columns of $\widehat{O}$, the rows and columns of $\widehat{T}$, and the rows of $\widehat{w}$) with two actions $(a,a')$ that are associated with that state. This is possible because the true observation matrix entries for the actions of a hidden state must be non-zero, and the true value of all other entries (for other actions) must be zero;
therefore, as long as we have sufficiently accurate estimates of the observation matrix, we can use the observation matrix parameters to augment the states $h$ with their associated action pair. This procedure is performed by Algorithm~\ref{alg:labelactions}. This labeling provides a connection between the HMM state $h$ and the original POMDP state. For a particular pair of actions $a,a'$, there are exactly $|S|$ HMM states that correspond to them. Thus looking at the columns of $\widehat{O}$ from left-to-right, and only picking out the columns that are labeled with $a,a'$ results in a specific ordering of the states $(a,\cdot,a')$, which is a permutation of the POMDP states, which we denote as $\{ s_{(a,a'),1}, s_{(a,a'),2}, \dots, s_{(a,a'),|S|} \}$. We will also use the notation $s_{(a,a')}$ to implicitly refer to a vector of states in the order of the permutation.

The algorithm proceeds to 
estimate the original POMDP parameters in order to perform 
planning and compute a policy. 
%of the original POMDP parameters in order to perform planning and compute a policy. 
Note that the estimated parameters use the computed $s_{(a,a')}$ 
permutations of the state. Let $\widehat{O}^{a,a'}$ be the submatrix where the rows and columns correspond to the actions $(a,a')$ and $\widehat{T}^{a,a',a''}$ be the submatrix where the rows correspond to the actions $(a',a'')$ and columns correspond to the actions $(a,a')$.
Then the estimated POMDP parameters can be computed 
as follows:  
\begin{align*}
\widehat{b}(s_{(a_0,a_1)}) &= normalize((\widehat{T}^{-1}\widehat{T}^{-1}\widehat{w})(a_0,\cdot,a_1)) \\
\widehat{p}(z | a, s_{(a,a')}) &= normalize(\sum_r \widehat{O}^{a,a'})\\
\widehat{p}(r | s_{(a,a')},a') &= normalize(\sum_z \widehat{O}^{a,a'}) \\
\widehat{p}(s_{(a',a'')} | s_{(a,a')}, a') &= normalize(\widehat{T}^{a,a',a''})
\end{align*}
Note that we require an additional $normalize()$ procedure 
since 
the MoM approach we leverage 
 is not guaranteed to return well formed probability distributions. 
The normalization procedure just divides by the sum to make them into valid probability distributions (if there are negative values we can either set them to zero or even just use the absolute value).

%We then use the 
% estimated POMDP parameters to compute a policy.
%As detailed in  
Algorithm \ref{alg:findpolicy} then uses these  
estimated POMDP parameters to compute a policy. 
The algorithm  constructs  $\beta$-vectors (see Definition \ref{def:beta}) 
that represent the expected sum of rewards of following a particular 
policy starting with action $a'$ given an input permuted state $s_{(a,a')}$. 
Aside from this slight modification, $\beta$-vectors 
are analogous to $\alpha$-vectors in standard POMDP planning. 
%which are the same as $\alpha$-vectors but they take as input the permuted states $s_{(a,a')}$ whose root action must match $a'$ (see Lemma TODO). 
The $\beta$-vectors form an approximate value function 
for the underlying POMDP and can be used in a similar 
way to standard $\alpha$-vectors.

\subsection{Phase 2}
In phase 2, after estimating the POMDP parameters and $\beta$-vectors, we use the estimated POMDP value function to 
extract a policy for acting, and we will shortly prove 
sufficient conditions for this policy to be near-optimal 
for all remaining episodes. 

The policy followed depends on the computed value function. 
If computationally tractable, one can compute $\beta$-vectors 
incrementally for all possible $H$-step policies. In 
this case, control proceeds by finding the best $\beta$-vector for the estimated initial belief $\widehat{b}(s_{(a_0,a_1)})$ (largest dot product of the $\beta$-vector with the initial belief) and then following the associated policy $\widehat{\pi}$. $\widehat{\pi}$ is then followed for the entire episode 
with no additional belief updating required as the policy itself  
 encodes the conditional branching.

However, in practical circumstances, it will not be 
possible to enumerate all possible $H$-step policies. 
In this case, one can use point-based approaches or 
other methods that use $\alpha$-vectors to enumerate only a subset of possible policies. 
In this case there will be an additional error $\epsilon_{planning}$ in the 
final error bound due to finite set of policies 
considered. In our analysis we omit $\epsilon_{planning}$ for simplicity and assume that 
we enumerate all $H$-step policies. %\textbf{**EB: is this correct?}. DG: yes it is

% are  computed incrementally from $t=1$ to $t=H$ 
%in the standard way for all possible $H$-step policies. Finally, 

\begin{definition}
\label{def:beta}
A $\beta$-vector taking as input $s_{(a,a')}$ with root action $a'$ and $t$-step conditional policies $f_t(r,z)$ for each observation pair $(r,z)$ is defined as
\begin{align*}
& \beta^{a'}_1(s_{(a,a')}) = \sum_r p(r|s_{(a,a')},a) \cdot r \\
& \beta_{t+1}^{a',f_t}(s_{(a,a')}) = \sum_{r,z,s_{(a',f_t(r,z))}} (r + \gamma \beta_{t}^{f_t(r,z)}(s_{(a',f_t(r,z))}))   \\
&\quad \cdot  p(r | s_{(a,a')}, a) p(z | s_{(a',f_t(r,z))}, a) p(s_{(a',f_t(r,z))} | s_{(a,a')}, a) 
\end{align*}
where $f_t(r,z)$ can also denote the root action of the policy $f_t(r,z)$ used in terms like $s_{(a,f_t(r,z))}$.
\end{definition}

\section{THEORY}

\subsection{PAC Theorem Setup}
We now state our primary result. For full details, please refer to our tech report\footnote{\url{http://www.cs.cmu.edu/~zguo/\#publications}}. Before doing so, we define some 
additional notation. 
Let $V^\pi(b) = \sum_{i=1}^H r_t$ starting from belief $b$ be the total undiscounted reward following policy $\pi$ for an episode. %Assume the initial belief $b$ is the same for all episodes.
Let $\sigma_{1,a}(T_a) = \max_a \sigma_1(T_a)$ and similarly for $\sigma_{1,a}(R_a)$ and $\sigma_{1,a}(Z_a)$. Let $\underline{\sigma}_{a}(T_a) = \min_a \sigma_{|S|}(T_a)$ and 
similarly for $\underline{\sigma}_{a}(R_a)$ and $\underline{\sigma}_{a}(Z_a)$.
Assume $\underline{\sigma}_{a}(T_a)$, $\underline{\sigma}_{a}(R_a)$, and $\underline{\sigma}_{a}(Z_a)$ are all at most $1$ (otherwise each term can be replaced by $1$ in the final sample complexity bound below).
\subsection{PAC Theorem}

\begin{theorem}
For POMDPs that satisfy the stated assumptions defined in the 
problem setting, executing \porl\ 
%Executing Algorithm \ref{alg:pac} 
will achieve an expected episodic reward of $V(b_0) \geq V^*(b_0) - \epsilon$ 
on all but a number of episodes that is bounded by 
\begin{eqnarray*}
O \left( \frac{H^4 V_{\max}^2|A|^{12} |R|^{4}|Z|^{4}|S|^{12}  \left(1+ \sqrt{\log(\frac{3}{\delta}})\right)^2 \log \left( \frac{3}{\delta} \right)}{ C_{d,d,d}\left(\frac{\delta}{3}^2\right) \underline{\sigma}_{a}(T_a)^{6} \underline{\sigma}_{a}(R_a)^8 \underline{\sigma}_{a}(Z_{a})^{8}  \epsilon^2}  \right)
\end{eqnarray*}
with probability at least $1-\delta$, where  
%bad episodes in which $V(b) \leq V^*(b) - \epsilon$ with probability $1-\delta$ where
\begin{align*}
C_{d,d,d}(\delta) &= \min(C_{1,2,3}(\delta),C_{1,3,2}(\delta)) \\
C_{1,2,3}(\delta) &= \min \left( \frac{\min_{i\neq j}||M_3(\vec{e}_i - \vec{e}_j)||_2  \cdot \sigma_k(P_{1,2})^2}{||P_{1,2,3}||_2 \cdot k^5 \cdot \kappa(M_1)^4} \right. \\
& \left. \cdot \frac{\delta}{\log (k / \delta)}, \frac{\sigma_k(P_{1,3})}{1} \right) \\
C_{1,3,2}(\delta) &= \min \left( \frac{\min_{i\neq j}||M_2(\vec{e}_i - \vec{e}_j)||_2  \cdot \sigma_k(P_{1,3})^2}{||P_{1,3,2}||_2 \cdot k^5 \cdot \kappa(M_1)^4}   \right. \\ 
& \left. \cdot \frac{\delta}{\log (k / \delta)} , \frac{\sigma_k(P_{1,2})}{1} \right)
\end{align*}
\end{theorem}
The quantities $C_{1,2,3},C_{1,3,2}$ directly arise from using
 the previously referenced MoM method for HMM parameter 
estimation~\citep{anandkumar2012method} and involve singular values of the moments of the induced HMM and the induced HMM parameters (see \citep{anandkumar2012method} for details).

We now briefly overview the proof. Detailed proofs are available in the supplemental material.
We first show that by executing \porl\, we obtain 
parameter estimates 
of the induced HMM, and bounds on these estimates, 
 as a function of the number of data points (Lemma~\ref{lem:mom}).  
We then prove that we can use the induced HMM to obtain estimated parameters of the underlying POMDP (Lemma \ref{lem:newquantities}). Then we show that we can compute policies that are equivalent (in structure and value) to those from the original POMDP (Lemma~\ref{lem:betaequiv}). We then bound the error in the resulting value function estimates of the resulting policies due to the use of approximate (instead of exact) model parameters (Lemma~\ref{lem:alphaerror}). 
%, as well as the error due to approximate POMDP planning (Lemma D). 
This allows us to compute a 
bound on the number of required samples (episodes) necessary to achieve near-optimal 
policies, with high probability, for use in phase 2. 

We commence the proof by bounding the error in estimates of the induced HMM parameters. In order to do that, 
we introduce Lemma \ref{lem:hmm}, which proves that samples taken in phase 1 
belong to an induced HMM where the transition and observation matrices are full rank. 
This is a requirement for being able to apply the MoM HMM parameter estimation 
procedure of \citet{anandkumar2012method}. 

\begin{lemma}
\label{lem:hmm}
The induced HMM has the observation and transition matrices defined as
\begin{align*}
& O(x_t^i,h_t^j) = \\
&\quad \delta(a_t^i, a_t^j)\delta(a_{t+1}^i,a_{t+1}^j)p(z_{t+1}^i | a_t^j, s_{t+1}^j)p(r_{t+1}^i | s_{t+1}^j, a_{t+1}^j) \\
& T(h_{t+1}^i,h_t^j) = \delta(a_{t+1}^i,a_{t+1}^j) p( s_{t+2}^i | s_{t+1}^j, a_{t+1}^j)p( a_{t+2}^i | a_t^j)
\end{align*}
where $i$ is the index over the rows and $j$ is the index over the columns, and $x_t^i = (a_t^i, z_{t+1}^i, r_{t+1}^i, a_{t+1}^i)$, $h_{t+1}^i=(a_{t+1}^i, s_{t+2}^i, a_{t+2}^i)$, $h_t^j=(a_t^j, s_{t+1}^j, a_{t+1}^j)$. $T$ and $O$ are both full rank and $w=p(h_2)$ has positive probability everywhere. Furthermore the following terms are bounded: $\Vert T \Vert_2 \leq \sqrt{|S|}$, $\Vert T^{-1} \Vert_2 \leq \frac{2(1+c|A|)}{\underline{\sigma}_{a}(T_a)}$, $\sigma_{\min}(O) \geq \underline{\sigma}_{a}(R_a)\underline{\sigma}_{a}(Z_{a})$, and $\Vert O \Vert_2 = \sigma_{1}(O) \leq |S|$.
\end{lemma}

Next, we use Lemma \ref{lem:mom}, which is an 
extension of the method of moments method by~\citet{anandkumar2012method} that provides a bound on 
the accuracies of the estimated induced HMM parameters in terms of $N$, the number of samples collected. Our extension involves computing $\widehat{T}$ (the original method only had $\widehat{O}$ and $\widehat{OT}$) and bounding its accuracy.

\begin{lemma}
\label{lem:mom}
Given an HMM such that $p(h_2)$ has positive probability everywhere, the transition matrix is full rank, and the observation matrix is full column rank, then by gathering $N$ samples of $(x_1, x_2, x_3)$, the estimates $\widehat{T},\widehat{O},\widehat{w}$ can be computed such that
\begin{align*}
||\widehat{T} - T||_2 &\leq   18 |A| |S|^{4} (\underline{\sigma}_{a}(R_a) \underline{\sigma}_{a}(Z_{a}))^{-4} \epsilon_1 \\
||\widehat{O} - O||_2 &\leq |A| |S|^{0.5} \epsilon_1\\
||\widehat{O} - O||_{\max} &\leq \epsilon_1\\
||\widehat{w} - w||_2 &\leq 14|A|^2|S|^{2.5} (\underline{\sigma}_{a}(R_a) \underline{\sigma}_{a}(Z_{a}))^{-4} \epsilon_1
\end{align*}
where $||\cdot||_2$ is the spectral norm for matrices, and the euclidean norm for vectors, and $w$ is the marginal probability of $h_2$, with probability $1-\delta$, as long as
\begin{align*}
N &\geq O \left( \frac{|A|^2 |Z| |R| (1+ \sqrt{\log(1/\delta)})^2}{(C_{d,d,d}(\delta))^2 \cdot \epsilon^2_1} \log \left( \frac{1}{\delta} \right) \right)
\end{align*}
\end{lemma}

Next we proceed by showing how to bound the error in the estimates of the POMDP parameters. The following Lemma \ref{lem:labelactions} is a prerequisite for computing the submatrices of $\widehat{O}$ and $\widehat{T}$ needed for the estimates of the POMDP parameters. 

\begin{lemma}
\label{lem:labelactions}
Given $\widehat{O}$ with max-norm error $\epsilon_O \leq \frac{1}{3|Z||R|}$, then the columns  which correspond to HMM states of the form $h=(a,s',a')$ can be labeled with their corresponding $a,a'$ using Algorithm \ref{alg:labelactions}.
\end{lemma}

With the correct labels, the submatrices of $\widehat{O}$ and $\widehat{T}$ allow us to compute estimates of the original POMDP parameters in terms of these permutations $s_{(a,a')}$.  Lemma \ref{lem:newquantities} bounds the error in these resulting estimates. 

\begin{lemma}
\label{lem:newquantities}
Given $\widehat{T},\widehat{O},\widehat{w}$ with max-norm 
errors $\epsilon_T,\epsilon_O,\epsilon_w$ respectively, then 
the following bounds hold on the estimated POMDP model parameters 
with probability at least $1-\delta$
\begin{align*}
&|\widehat{p}(s_{(a',a'')} | s_{(a,a')}, a') - p(s_{(a',a'')} | s_{(a,a')}, a')| \leq \frac{4|S| \epsilon_T}{\epsilon_a^2} \\
& |\widehat{p}(z | a, s_{(a,a')}) - p(z | a, s_{(a,a')})| \leq 4|Z||R|\epsilon_O \\
& |\widehat{p}(r | s_{(a,a')},a') - p(r | s_{(a,a')},a')|  \leq 4|Z||R|\epsilon_O \\
& |\widehat{b}(s_{(a_0,a_1)}) - b(s_{(a_0,a_1)})| \\
&\quad \leq 4|A|^4|S|(||T^{-1}||_2^2 \epsilon_w + 6 ||T^{-1}||_2^3 \epsilon_T)
\end{align*}
where $\epsilon_a = \Theta(1/|A|)$
\end{lemma}

We proceed by 
bounding the error in computing the estimated $\beta$-vectors. 
Lemma \ref{lem:betaequiv} states that $\beta$-vectors are equivalent under permutation to $\alpha$-vectors.

\begin{lemma}
\label{lem:betaequiv}
Given the permutation of the states $s_{(a,a'),j} = s_{\phi((a,a'),j)}$,
$\beta$-vectors and $\alpha$-vectors over the same policy $\pi_t$ are equivalent i.e. $\beta_t^{\pi_t}(s_{(a,a'),j}) = \alpha^{\pi_t}_t(s_{\phi((a,a'),j)})$
\end{lemma}

The following lemma bounds the error in the resulting 
$\alpha$-vectors obtained by performing POMDP planning, 
and follows from prior work~\citep{fard2008variance,ross2009sensitivity}.
\begin{lemma}
\label{lem:alphaerror}
Suppose we have approximate POMDP parameters with errors $|\widehat{p}(s' | s, a) - p(s' | s, a)| \leq \epsilon_T$, $|\widehat{p}(z | a, s') - p(z | a, s')| \leq \epsilon_Z$, and $|\widehat{p}(r | s,a) - p(r | s,a')| \leq \epsilon_R$. Then for any $t$-step conditional policy  $\pi_t$
\begin{eqnarray*}
|\alpha^{\pi_t}_{t}(s) - \widehat{\alpha}^{\pi_t}_{t}(s)| &\leq t^2 R_{\max} (|R|\epsilon_R +  |S|\epsilon_T + |Z|\epsilon_Z ).
\end{eqnarray*}
\end{lemma}

We next prove that our \porl\ algorithm computes a policy that is optimal 
for the input parameters\footnote{Again, we could easily modify this 
to account for approximate planning error, but leave this out for simplicity, as 
we do not expect this to make a significant impact on the resulting sample 
complexity, except in terms of minor changes to the polynomial terms.}: 
\begin{lemma}
\label{lem:findpolicy}
Algorithm \ref{alg:findpolicy} finds the policy $\widehat{\pi}$ which maximizes $V^{\widehat{\pi}}(\widehat{b}(s_1))$ for a POMDP with parameters
%\begin{eqnarray*}
$\widehat{b}(s_1), \widehat{p}(z | a, s'), \widehat{p}(r | s,a),$ and $\widehat{p}(s' | s, a)$.
%\end{eqnarray*}
\end{lemma}

We now have all the key pieces to prove our result. 

\begin{proof}
(Proof sketch of Theorem 1).  Lemma \ref{lem:newquantities} shows that the error in the estimates of the POMDP parameters can be bounded in terms of the error in the induced HMM parameters, 
which is itself bounded in terms of the number of samples (Lemma~\ref{lem:hmm}).
Lemma~\ref{lem:betaequiv} and Lemma \ref{lem:alphaerror}
 together  bound in the error in computing the estimated value function 
(as represented by $\beta$-vectors)  
using estimated POMDP parameters.

We then need to bound the error from executing $\widehat{\pi}$ that Algorithm \ref{alg:findpolicy} returns compared to the optimal policy $\pi^*$. We know from Lemma \ref{lem:findpolicy} that Algorithm \ref{alg:findpolicy} correctly identifies the best policy for the estimated POMDP. Then let the initial beliefs $b,\widehat{b}$ have error $\Vert b - \widehat{b} \Vert_{\infty} \leq \epsilon_b$, and the bound over $\alpha$-vectors of any policy $\pi$, $\Vert \alpha^{\pi} - \widehat{\alpha}^{\pi} \Vert_{\infty} \leq \epsilon_{\alpha}$ be given. Then
\begin{align*}
\widehat{V}^{\widehat{\pi}}(\widehat{b}) &= \widehat{b} \cdot \widehat{\alpha}^{\widehat{\pi}} \geq \widehat{b} \cdot \widehat{\alpha}^{\pi^*} \\
&\geq \widehat{b} \cdot \alpha^{\pi^*} - |\widehat{b} \cdot \alpha^{\pi^*} -  \widehat{b} \cdot \widehat{\alpha}^{\pi^*}| \geq \widehat{b} \cdot \alpha^{\pi^*} - \epsilon_{\alpha} \\
&\geq b \cdot \alpha^{\pi^*} - |b \cdot \alpha^{\pi^*} - \widehat{b} \cdot \alpha^{\pi^*}| - \epsilon_{\alpha} \\
&\geq b \cdot \alpha^{\pi^*} - \epsilon_{b} V_{\max} - \epsilon_{\alpha} = V^*(b) - \epsilon_{b} V_{\max} - \epsilon_{\alpha}
\end{align*}
where the first inequality is because $\widehat{\pi}$ is the optimal policy for $\widehat{b}$ and $\widehat{\alpha}$, the second inequality is by the triangle inequality, the third inequality is because $\Vert \widehat{b} \Vert_1 = 1$, the fourth inequality is by the triangle inequality, the fifth inequality is since $\alpha$ is at most $V_{\max}$. Next
\begin{align*}
V^{\widehat{\pi}}(b) &= b \cdot \alpha^{\widehat{\pi}} \geq \widehat{b} \cdot \alpha^{\widehat{\pi}} - |\widehat{b} \cdot \alpha^{\widehat{\pi}} - b \cdot \alpha^{\widehat{\pi}}| \\
& \geq \widehat{b} \cdot \alpha^{\widehat{\pi}} - \epsilon_{b}V_{\max} \\
&\geq \widehat{b} \cdot \widehat{\alpha}^{\widehat{\pi}} - |\widehat{b} \cdot \widehat{\alpha}^{\widehat{\pi}} - \widehat{b} \cdot \alpha^{\widehat{\pi}}| - \epsilon_{b}V_{\max} \\
&\geq \widehat{b} \cdot \widehat{\alpha}^{\widehat{\pi}} - \epsilon_{\alpha} - \epsilon_{b}V_{\max}
\end{align*}
where the first inequality is by triangle inequality, the second inequality is because $\alpha$ is at most $V_{\max}$, the third inequality is triangle inequality, and the fourth inequality is due to $\Vert \widehat{b} \Vert_1 = 1$. Putting those two together results in
\begin{eqnarray*}
V^{\widehat{\pi}}(b) &\geq V^*(b) - 2\epsilon_{b} V_{\max} - 2\epsilon_{\alpha}
\end{eqnarray*}
Letting $\epsilon = 2\epsilon_{b} V_{\max} + 2\epsilon_{\alpha}$, and setting the 
number of episodes $N$ to the value specified in the theorem will ensure that 
the resulting 
errors $\epsilon_{b}$ and $\epsilon_{\alpha}$ are small enough to obtain an $\epsilon$-optimal 
policy as desired. 
\end{proof}
%with the bounds from propagating the Lemmas will result in the final bound in the main PAC theorem.

	\section{CONCLUSION}
	
	We have provided a PAC RL algorithm for an important class of episodic POMDPs, which includes many information gathering domains. To our knowledge this is the first RL algorithm for partially observable settings that has a sample complexity that is a polynomial function of the POMDP parameters.
	
	There are many areas for future work. We are interested in reducing the set of currently required assumptions, thereby creating PAC PORL algorithms that are suitable to more generic settings. Such a direction may also require exploring alternatives to method of moments approaches for performing latent variable estimation. We also hope that our theoretical results will lead to further insights on practical algorithms for partially observable RL.
	
	{
		\small
		\subsubsection*{Acknowledgements}
		This work was supported by NSF CAREER grant 1350984.
	}
	\newpage
	
	{
		\small
		%\bibliographystyle{plainnat}
		%\bibliography{porl}

	}
	
	% appendix
	{
	\newpage
	\onecolumn
	\renewcommand\thesection{\Alph{section}}
	\setcounter{section}{0}
	\setcounter{algocf}{0}
	\section{Appendix Overview}

This appendix is organized as follows. Section \ref{sec:apphelpers}, contains a few generic helper lemmas. Section \ref{sec:appalgorithm} contains pseudo-code of the algorithm. Section \ref{sec:app:small} contains some small lemmas to be used later. Section \ref{sec:app:main} states the main lemmas, which are also stated in the main paper. Finally, section \ref{sec:app:theorem} contains the main theorem and proof of the paper.

\section{Helper Lemmas}
\label{sec:apphelpers}

\subsection{Matrix Norms Lemma}

Taken from the matrix norms section in the matrix cookbook \cite{petersen2008matrix}.

\begin{enumerate}
	\item Induced norms (e.g. $||\cdot||_2$) are sub-multiplicative: $||AB||_2 \leq ||A||_2 ||B||_2$
	\item $||A||_2 \leq ||A||_F$
	\item $||A||_{\max} \leq ||A||_2$ where $||A||_{\max} = \max_{ij}|A_{ij}|$
	\item $||A||_{\max} \leq ||A||_F$
	\item $||A||_F \leq \sqrt{d}||A||_2$ where $A$ has rank $d$
	\item $||A||_F \leq \sqrt{mn}||A||_{\max}$ where $A$ is $m \times n$
	\item $||A||_2 \leq \sqrt{mn}||A||_{\max}$
	\item $||AB||_{\max} \leq \sqrt{mn^2 k}||A||_{\max}||B||_{\max}$ derived from $||\cdot||_2$ where $B$ is $n \times k$
\end{enumerate}

\subsection{Perturbed Inverse Lemma}

Taken from MoM paper \cite{anandkumar2012method}, which was taken from Theorem 2.5, p. 118 in \cite{stewart1990matrix}.

\emph{If}
\begin{enumerate}
	\item $A,E \in \mathbb{R}^{k \times k}$, $A$ is invertible
	\item $||A^{-1} E ||_2 < 1$ ($||A^{-1}||_2||E||_2 < 1$ is sufficient)
	\item $\widehat{A} = A + E$
\end{enumerate}
\emph{Then}
\begin{enumerate}
	\item $\widehat{A}$ is invertible
	\item 
	\begin{align}
	||\widehat{A}^{-1} - A^{-1}||_2 \leq \frac{||E||_2 ||A^{-1}||_2^2}{1 - ||A^{-1}E||_2} 
	\leq \frac{||E||_2 ||A^{-1}||_2^2}{1 - ||A^{-1}||_2 ||E||_2}
	\end{align}
\end{enumerate}

\subsection{Submatrix Eigenvalue Extreme Lemma}

Let $M$ be a $n\times n$ symmetric matrix, which can be viewed as a block matrix
\begin{align}
M &= \left[ \begin{matrix}
A & B \\
C & D
\end{matrix} \right]
\end{align}
where $A$ is $k \times k$ and $B,C,D$ have appropriate dimensions. Then $\lambda_n(M) \leq \lambda_k(A) \leq \lambda_1(A) \leq \lambda_1(M)$ i.e. the min and max eigenvalues of $A$ are bounded in-between the min and max eigenvalues of $M$.

\begin{proof}
The Rayleigh quotient with $M$ is
\begin{align}
R(M,x) &= \frac{x^T M x}{x^T x}
\end{align}
such that $\min_x R(M,x)$ is equal to the smallest eigenvalue $\lambda_n(M)$ and $\max_x R(M,x)$ is equal to the largest eigenvalue $\lambda_1(M)$. Let $x^T = \left[ y^T z^T \right]$ where $y$ is column vector of size $k$. Note that $A$ is symmetric as well. Then
\begin{align}
R(M,x) &= \frac{\left[ y^T z^T \right] \left[ \begin{matrix}
	A & B \\
	C & D
	\end{matrix} \right] \left[ \begin{matrix}
	y \\
	z
	\end{matrix} \right]}{\left[ y^T z^T \right] \left[ \begin{matrix}
	y \\
	z
	\end{matrix} \right]} \\
&= \frac{y^T A y + z^T C y + y^T B z + z^T D z}{ y^T y + z^T z} \\
\implies R(M, [y^T \; 0]) &= \frac{y^T A y}{ y^T y} \\
&= R(A,y)
\end{align}
This means that $\min_x R(M,x) \leq \min_y R(M, [y^T \; 0]) = \min_y R(A,y) \leq \max_y R(A,y) = \max_y R(M, [y^T \; 0]) \leq \max_x R(M,x)$. Thus $\lambda_n(M) \leq \lambda_k(A) \leq \lambda_1(A) \leq \lambda_1(M)$.
\end{proof}

\subsection{Normalization Lemma}

Given a nonnegative, nonzero vector $x \in \mathbb{R}^n$ and an estimated vector $\tilde{x} \in \mathbb{R}^n$ such that $\Vert x - \tilde{x}\Vert_{\infty} \leq \epsilon \leq \frac{\Vert x \Vert_1}{2n}$, let the normalization function be $\widehat{x} = normalize(\tilde{x})$ where $\widehat{x}_i = \frac{|\tilde{x}_i|}{\Vert \tilde{x} \Vert_1}$. Then
\begin{align}
\Vert normalize(x) - normalize(\tilde{x})  \Vert_{\infty} &\leq \frac{2(x_i n + \Vert x \Vert_1) \epsilon}{\Vert x \Vert_1^2}
\end{align}

\begin{proof}

Let $abs(\tilde{x})_i = |\tilde{x}_i|$. Then $\left| abs(\tilde{x})_i - x_i \right| = \left| |\tilde{x}_i| - x_i \right| \leq \left| \tilde{x}_i - x_i \right|$ since $x_i \geq 0$. This implies that $\Vert abs(\tilde{x}) - x \Vert_{\infty} \leq \Vert x - \tilde{x}\Vert_{\infty} \leq \epsilon$. Next note that $ \left| (\sum_i x_i) - \sum_i (abs(\tilde{x})_i)  \right| \leq \sum_i \left| x_i - abs(\tilde{x})_i  \right| $ which means $| \Vert x \Vert_1 - \Vert abs(\tilde{x}) \Vert_1 | \leq \Vert x - abs(\tilde{x}) \Vert_1 \leq n\epsilon$. Note that $\Vert abs(\tilde{x})\Vert_1 = \Vert \tilde{x} \Vert_1$. Then
\begin{align}
\left| normalize(x)_i - normalize(\tilde{x})_i \right| &= \left| \frac{x_i}{\Vert x \Vert_1} - \frac{|\tilde{x}_i|}{\Vert \tilde{x} \Vert_1} \right| \\
&\leq  \frac{\left| x_i \Vert \tilde{x} \Vert_1 - |\tilde{x}_i| \Vert x \Vert_1 \right|}{\Vert x \Vert_1 \Vert \tilde{x} \Vert_1} \\
&\leq  \frac{\left| x_i \Vert \tilde{x} \Vert_1 - x_i \Vert x \Vert_1 \right| + \left| x_i \Vert x \Vert_1 - |\tilde{x}_i| \Vert x \Vert_1 \right|}{\Vert x \Vert_1 (\Vert x \Vert_1 - n\epsilon)} \\
&\leq  \frac{x_i n\epsilon + \Vert x \Vert_1 \epsilon}{\frac{1}{2} \Vert x \Vert_1^2} \\
&\leq  \frac{2(x_i n + \Vert x \Vert_1) \epsilon}{\Vert x \Vert_1^2}
\end{align}
where the numerator of the third inequality follows from $\Vert x - abs(\tilde{x}) \Vert_1 \leq n\epsilon$ that was derived above.
\end{proof}

\section{Algorithm}
\label{sec:appalgorithm}

\begin{algorithm}[H]
	\SetKwInOut{Input}{input}
	\Input{$S,A,Z,R,H,N,c,\pi_{rest}$}
	Let $\pi_{explore}$ be the policy where $a_1,a_2$ are uniformly random, and $p(a_{t+2}|a_t) = \frac{1}{1 + c|A|}(\mathbf{I} + c\mathbf{1}_{|A|\times|A|})$ \;
	$X \gets \emptyset$ \;
	\tcp{Phase 1:}
	\tcp{Collect samples for an induced HMM}
	\For{episode $i \gets 1$ \KwTo $N$}{
		Follow $\pi_{explore}$ for 4 steps \;
		Let $x_t = (a_t, r_t,z_t,a_{t+1})$ \;
		\tcp{collect observation sample tuple}
		$X \gets X \cup \{(x_1, x_2, x_3)\}$ \;
		Execute $\pi_{rest}$ for the rest of the steps \;
	}
	\tcp{\textbf{Phase 2:}}
	\tcp{Compute estimated POMDP parameters}
	Get $\widehat{T},\widehat{O},\widehat{w}$ for the induced $HMM$ from $X$ through an adapted MoM method \;
	Call Algorithm \ref{alg:app:labelactions} with $\widehat{O}$ to label the columns with their corresponding actions \;
	Let $s_{(a,a'),i}$ be a permutation of states by following the ordering of the columns of $\widehat{O}$ corresponding to $h=(a,\cdot,a')$ \;
	Let $\widehat{O}^{a,a'}$ be the submatrix where the rows and columns correspond to the actions $(a,a')$ \;
	Let $\widehat{T}^{a,a',a''}$ be the submatrix where the rows correspond to the actions $(a',a'')$ and columns correspond to the actions $(a,a')$ \;
	Compute the following estimates of the POMDP parameters:
	\begin{align*}
	\widehat{b}(s_{(a_0,a_1)}) = normalize((\widehat{T}^{-1}\widehat{T}^{-1}\widehat{w})(a_0,\cdot,a_1)) \\
	\widehat{p}(z | a, s_{(a,a')}) = normalize(\sum_r \widehat{O}^{a,a'})\\
	\widehat{p}(r | s_{(a,a')},a') = normalize(\sum_z \widehat{O}^{a,a'})\\
	\widehat{p}(s_{(a',a'')} | s_{(a,a')}, a') = normalize(\widehat{T}^{a,a',a''})
	\end{align*}
	Call Algorithm \ref{app:alg:findpolicy} with the above quantities to estimate a policy $\widehat{\pi}$ \;
	Execute $\widehat{\pi}$ for the rest of the episodes \;
	
	\caption{\porl}
	\label{alg:app:pac}
\end{algorithm}

\begin{algorithm}[h]
	\SetKwInOut{Input}{input}
	\Input{$\widehat{O}$}
	\ForEach{column $i$ of $\widehat{O}$}{
		Find a row $j$ such that $\widehat{O}(i,j) \geq 
		\frac{2}{3|R||Z|}$ \;
		Let the observation associated with row $j$ be $(a,r',z',a')$, then label column $i$ with $(a,a')$ \;
	}
	\caption{LabelActions}
	\label{alg:app:labelactions}
\end{algorithm}

\begin{algorithm}[h]
	\SetKwInOut{Input}{input}
	\Input{
		\begin{align*}
		& \widehat{b}(s_{(a_0,a_1)})\\
		& \widehat{p}(z | a, s_{(a,a')})\\
		&\widehat{p}(r | s_{(a,a')},a')\\
		&\widehat{p}(s_{(a',a'')} | s_{(a,a')}, a')
		\end{align*}
	}
	\tcp{First, incrementally compute $\beta$-vectors for all $t$-step policies}
	\tcp{$\Gamma^{a_-,a}_t$ contains all $\beta$-vectors for $t$-step policies that take as input the permtuation $s_{(a_-,a)}$}
	$\forall a_-,a \in A, \quad \Gamma^{a_-,a}_1 = \{ \widehat{\beta}_1^a(s_{(a_-,a)}) \}$ \;
	\For{$t \gets 2$ \KwTo $H$}{
		$\forall a,a' \in A, \Gamma^{a,a'}_t = \emptyset$ \;
		\For{$a,a' \in A$}
		{
			\For{$f_t(r,z) \in (|R|\times|Z| \rightarrow \Gamma_{t-1}^{a,a'})$}{
				\tcp{all mappings from an observation pair to a previous $\beta$-vector}
				$\forall a_- \in A, \quad \Gamma^{a_-,a}_t = \Gamma^{a_-,a}_t \cup \{ \beta_t^{a,f_t}(s_{(a_-,a)}) \}$ \;
			}
		}
	}
	Return $\arg\max_{a_0,a_1,\beta_H(s_{(a_0,a_1)})) \in\Gamma_H^{a_0,a_1}} \sum_{s_{(a_0,a_1)}} \widehat{b}(s_{(a_0,a_1)}) \cdot \beta_H(s_{(a_0,a_1)})$  \;
	\caption{FindPolicy}
	\label{app:alg:findpolicy}
\end{algorithm}

\section{Small Lemmas}
\label{sec:app:small}

\subsection{Spectral Norm Lemma}

\begin{applemma*}
	\label{lem:app:probmatrix}
	For any $ a\in A$
	\begin{align}
	||T_a||_2 &\leq \sqrt{|S|} \\
	||Z_a||_2 &\leq \sqrt{|S|} \\
	||R_a||_2 &\leq \sqrt{|S|}
	\end{align}
\end{applemma*}

\begin{proof}
	
Using $||A||_2 \leq ||A||_F = ||vec(A)||_2 \leq \sqrt{||vec(A)||_1}$ where $vec(A)$ is the vector of the entries of the matrix $A$, and the last inequality holds because all entries of $A$ are at most 1
\begin{align}
||T_a||_2 &\leq \sqrt{|S|} \\
||Z_a||_2 &\leq \sqrt{|S|} \\
||R_a||_2 &\leq \sqrt{|S|}
\end{align}

\end{proof}

\subsection{Exploration Policy Lemma}

\begin{applemma*}
	\label{lem:app:explorationpolicy}
	Let $\Pi$ be the $|A| \times |A|$ matrix that $\pi_{explore}$ is based on where
	\begin{align}
	p(a_{t+2}| a_t) &= \Pi(a_{t+2},a_t) \\
	&= \frac{1}{1 + c|A|}(\mathbf{I} + c\mathbf{1}_{|A|\times|A|})
	\end{align}
	and $c>0$ is an open parameter, and $\mathbf{1}_{|A|\times|A|}$ is a $|A| \times |A|$ matrix of all ones. Then it follows that
	\begin{align}
	||\Pi||_2 &\leq 1 \\
	||\Pi^{-1}||_2 &\leq 2(1+c|A|) \\
	\epsilon_a &= \min_{a'',a} \Pi(a'',a) = \frac{c}{1+c|A|}
	\end{align}
\end{applemma*}

\begin{proof}

$\Pi$ is a perturbed identity matrix
\begin{align}
\Pi &= \frac{1}{1 + c|A|}(\mathbf{I} + c\mathbf{1}_{|A|\times|A|})
\end{align}
where $c > 0$ is a real number we can choose, and $\mathbf{1}_{|A|\times|A|}$ is a $|A| \times |A|$ matrix of all ones. This yields $\Pi$ such that each column is a probability distribution, where the off diagonal entries are all equal, and the diagonal entries are equal and greater than the off diagonal entries. Then
\begin{align}
\epsilon_a &= \min_{a'',a} \Pi(a'',a) = \frac{c}{1+c|A|}
\end{align}

Since $\mathbf{1}_{|A|\times|A|}$ has only a rank of 1, it can have at most one nonzero eigenvalue. Note $\mathbf{1}_{|A|\times|A|}  \mathbf{1}_{|A|} = |A|\mathbf{1}_{|A|}$ where $\mathbf{1}_{|A|}$ is a column vector of ones. Then $\mathbf{1}_{|A|\times|A|}$ has only one nonzero eigenvalue of $|A|$. Thus $||\mathbf{1}_{|A|\times|A|}||_2 = |A|$. Then
\begin{align}
\Vert \Pi \Vert_2 &= \left\Vert \frac{1}{1 + c|A|}(\mathbf{I} + c\mathbf{1}_{|A|\times|A|}) \right\Vert_2 \\
&\leq \frac{1}{1 + c|A|} (\Vert \mathbf{I} \Vert_2 + c \Vert \mathbf{1}_{|A|\times|A|} \Vert_2) \\
&\leq \frac{1}{1 + c|A|}(1 + c|A|) \\
&\leq 1
\end{align}
Next, by the Sherman-Morrison formula
\begin{align}
(\mathbf{I} + c\mathbf{1}_{|A|\times|A|})^{-1} &= (\mathbf{I} + c\mathbf{1}_{|A|}\mathbf{1}_{|A|}^T)^{-1} \\
&= \mathbf{I} - \frac{\mathbf{I} (c\mathbf{1}_{|A|})\mathbf{1}_{|A|}^T \mathbf{I}}{1 + \mathbf{1}_{|A|}^T \mathbf{I}(c\mathbf{1}_{|A|})} \\
&= \mathbf{I} - \frac{c\mathbf{1}_{|A|\times|A|}}{1 + c |A|}
\end{align}
then
\begin{align}
\Vert \Pi^{-1} \Vert_2 &= \left\Vert \left(\frac{1}{1 + c|A|}(\mathbf{I} + c\mathbf{1}_{|A|\times|A|}) \right)^{-1} \right\Vert_2 \\
&= (1 + c|A|) \left\Vert \left(\mathbf{I} + c\mathbf{1}_{|A|\times|A|}\right)^{-1} \right\Vert_2 \\
&= (1 + c|A|) \left\Vert \left(\mathbf{I} - \frac{c\mathbf{1}_{|A|\times|A|}}{1 + c |A|}\right) \right\Vert_2 \\
&\leq (1 + c|A|) \left( \left\Vert \mathbf{I} \right\Vert_2 + \frac{c}{1 + c |A|}\left\Vert \mathbf{1}_{|A|\times|A|} \right\Vert_2 \right) \\
&= (1 + c|A|) \left( 1 + \frac{c|A|}{1 + c |A|} \right) \\
&\leq (1 + c|A|) \left( 1 + 1 \right) \\
&= 2(1 + c|A|)
\end{align}
\end{proof}

\section{Main Lemmas}
\label{sec:app:main}

\subsection{Lemma \ref{lem:app:hmm} and Proof}
\begin{applemma}
	\label{lem:app:hmm}
	The induced HMM has the observation and transition matrices defined as
	\begin{align*}
	& O(x_t^i,h_t^j) = \\
	&\quad \delta(a_t^i, a_t^j)\delta(a_{t+1}^i,a_{t+1}^j)p(z_{t+1}^i | a_t^j, s_{t+1}^j)p(r_{t+1}^i | s_{t+1}^j, a_{t+1}^j) \\
	& T(h_{t+1}^i,h_t^j) =\\
	&\quad \delta(a_{t+1}^i,a_{t+1}^j) p( s_{t+2}^i | s_{t+1}^j, a_{t+1}^j)p( a_{t+2}^i | a_t^j)
	\end{align*}
	where $i$ is the index over the rows and $j$ is the index over the columns, and $x_t^i = (a_t^i, z_{t+1}^i, r_{t+1}^i, a_{t+1}^i),h_{t+1}^i=(a_{t+1}^i, s_{t+2}^i, a_{t+2}^i),h_t^j=(a_t^j, s_{t+1}^j, a_{t+1}^j)$. $T$ and $O$ are both full rank and $w=p(h_2)$ has positive probability everywhere. Furthermore
	\begin{eqnarray}
		\Vert T \Vert_2 &\leq \sqrt{|S|} \\
		\Vert T^{-1} \Vert_2 &\leq \frac{2(1+c|A|)}{\underline{\sigma}_{a}(T_a)} \\
		\sigma_{\min}(O) &\geq \underline{\sigma}_{a}(R_a)\underline{\sigma}_{a}(Z_{a}) \\
		\Vert O \Vert_2 &= \sigma_{1}(O) \leq |S|
	\end{eqnarray}
\end{applemma}

\emph{Proof:}

First we will show that the Markov property holds for the newly defined hidden state. Then we will show that the middle belief has positive probability everywhere. Then we will show that the observation and transition matrices of this induced HMM are as stated and have full column rank and have bounded singular values.

\subsubsection{HMM}

The HMM state is defined as $h_t = (a_t, s_{t+1}, a_{t+1})$, i.e. a state along with the previous and current actions. The new HMM observation will be $x_t = (a_t, z_{t+1}, r_{t+1}, a_{t+1})$. Note that the action is defined to only depend on the action 2 steps ago i.e. $p(a_{t+2} | a_t)$ and is independent of everything else. First, the Markov property will be shown to hold
\begin{align}
	p(h_{t+1} | \dots, h_t) &= p((a_{t+1}', s_{t+2}', a_{t+2}') | \dots, (a_t, s_{t+1}, a_{t+1})) \\
	&= p(a_{t+2}' | a_t) p(a_{t+1}' | a_{t+1}) p(s_{t+2}' | s_{t+1}, a_{t+1}) \\
	&= p(h_{t+1} | h_t)
\end{align}
Note that $p(a'_{t+1} | a_{t+1})$ is just a delta function.
Next, the HMM observation will be shown to only directly depend on the current HMM state
\begin{align}
	p(x_t | h_t, \dots) &= p(a_t', z_{t+1}, r_{t+1}, a_{t+1}' | a_t, s_{t+1}, a_{t+1}, \dots) \\
	&= p(a_t' | a_t)p(a_{t+1}' | a_{t+1}) p(z_{t+1} | a_t, s_{t+1}) p(r_{t+1} | s_{t+1}, a_{t+1}) \\
	&= p(x_t | h_t)
\end{align}
where $p(a_t' | a_t)$ and $p(a_{t+1}' | a_{t+1})$ are essentially delta functions.
Thus this formulation gives rise to an HMM.

\subsubsection{Middle Belief}

By the assumptions on the initial belief of the POMDP, after any $a_2$, all states are reachable for $s_3$. This means $p(a_2,s_3)$ has positive probability for all values. Furthermore, $a_3$ only depends on $a_1$. Since $a_1$ is uniformly random, by the choice of $\Pi$, $a_3$ is also uniformly random. Thus $p(a_2,s_3,a_3) = p(h_2)$ has positive probability for all values.

\subsubsection{Observation Matrix}

Next, the HMM observation matrix will be derived
\begin{align}
	O(x_t^i,h_t^j) &= p(a_t^i, z_{t+1}^i, r_{t+1}^i, a_{t+1}^i | a_t^j, s_{t+1}^j, a_{t+1}^j) \\
	&= \delta(a_t^i, a_t^j)\delta(a_{t+1}^i,a_{t+1}^j)p(z_{t+1}^i | a_t^j, s_{t+1}^j)p(r_{t+1}^i | s_{t+1}^j, a_{t+1}^j) \label{eq:app:hmmobs}
\end{align}
where $i$ is the index over the rows and $j$ is the index over the columns, and $x_t^i = (a_t^i, z_{t+1}^i, r_{t+1}^i, a_{t+1}^i),h_t^j=(a_t^j, s_{t+1}^j, a_{t+1}^j)$. Also note that the $t$ here doesn't really matter (since the matrix is the same for all $t$), and is only used in a relative way to distinguish between current and next time steps.

The next claim is that $O$ is full column rank. To see this, first permute the columns and rows of $O$ so that they are grouped by $(a_t^i,a_{t+1}^i)$ for the rows and $(a_t^j,a_{t+1}^j)$ for the columns. Then the only nonzero blocks are when $(a_t^i,a_{t+1}^i) = (a_t^j,a_{t+1}^j)$ i.e. $O$ becomes a block diagonal matrix with rectangular blocks. Next, fix $(a_t, a_{t+1}) = (a_t^i,a_{t+1}^i) = (a_t^j,a_{t+1}^j)$. Then the block corresponding to $(a_t, a_{t+1})$ is made up of the entries $p(z_{t+1}^i | a_t, s_{t+1}^j)p(r_{t+1}^i | s_{t+1}^j, a_{t+1})$. Call this block $O^{a_t,a_{t+1}}$. Consider looking only at the rows where we fix $r_{t+1}^i=r$, then the columns are just a scaled version of the columns of $Z_{a_t}$. Since $Z_{a_t}$ has full column rank, this implies that this block also has linearly independent columns. Since all of the diagonal blocks of $O$ have full column rank, this implies $O$ itself has full column rank.

Finally, consider the singular values of $O$. Recall $O$ can be viewed as a block diagonal matrix with rectangular blocks $O^{a_t,a_{t+1}}$. Then $(O^T O)$ is a block diagonal matrix with square blocks of $((O^{a_t,a_{t+1}})^T O^{a_t,a_{t+1}})$ on the diagonal. The eigenvalues of $(O^T O)$ are therefore just the eigenvalues of $((O^{a_t,a_{t+1}})^T O^{a_t,a_{t+1}})$ i.e. the singular values of $O$ are just the singular values of $O^{a_t,a_{t+1}}$. Note that $O^{a_t,a_{t+1}}((z,r),s) = p(z | a_t, s)p(r | s, a_{t+1})$ and can be thought of as the kronecker product $(R_{a_{t+1}} \otimes Z_{a_t})((r,z),(s_1,s_2)) = p(z | a_t, s_1)p(r | s_2, a_{t+1})$ but with all the columns removed except for the columns in which $s_1 = s_2 = s$. In other words the kronecker product can be thought of as the following block matrix
\begin{align}
	R_{a_{t+1}} \otimes Z_{a_t} &= \left[ O^{a_t,a_{t+1}} \quad B \right]
\end{align}
where $B$ is a matrix containing the columns of the kronecker product not present in $O^{a_t,a_{t+1}}$. Then
\begin{align}
	(R_{a_{t+1}} \otimes Z_{a_t})^T (R_{a_{t+1}} \otimes Z_{a_t}) &= \left[ \begin{matrix}
		((O^{a_t,a_{t+1}})^T O^{a_t,a_{t+1}}) & \dots \\
		\dots & \dots
	\end{matrix} \right]
\end{align}
Then by the submatrix eigenvalue extreme lemma, it follows that $\sigma_{\min}(R_{a_{t+1}} \otimes Z_{a_t}) \leq \sigma_{\min}(O^{a_t,a_{t+1}}) \leq \sigma_1(O^{a_t,a_{t+1}}) \leq \sigma_1(R_{a_{t+1}} \otimes Z_{a_t})$. Then
\begin{align}
	\sigma_{\min}(O) &= \sigma_{\min}(O^{a_t,a_{t+1}}) \\
	&\geq (\min_{a,a'}\sigma_{\min}(R_a)\sigma_{\min}(Z_{a'})) \\
	&\geq \underline{\sigma}_{a}(R_a)\underline{\sigma}_{a}(Z_{a}) \\
	\sigma_{1}(O) &= \sigma_{1}(O^{a_t,a_{t+1}}) \\
	&\leq (\max_{a,a'}\sigma_{1}(R_a)\sigma_{1}(Z_{a'})) \\
	&\leq |S|
\end{align}
where the last inequality follows from Lemma \ref{lem:app:probmatrix}.

\subsubsection{Transition Matrix}

Next, the HMM transition matrix will be derived
\begin{align}
	T(h_{t+1}^i,h_t^j) &= p( a_{t+1}^i, s_{t+2}^i, a_{t+2}^i | a_t^j, s_{t+1}^j, a_{t+1}^j) \\
	&= \delta(a_{t+1}^i,a_{t+1}^j) p( s_{t+2}^i, a_{t+2}^i | a_t^j, s_{t+1}^j, a_{t+1}^j) \\
	&= \delta(a_{t+1}^i,a_{t+1}^j) p( s_{t+2}^i | s_{t+1}^j, a_{t+1}^j)p( a_{t+2}^i | a_t^j)
\end{align}
where $i$ is the index over the rows and $j$ is the index over the columns, and $h_{t+1}^i=(a_{t+1}^i, s_{t+2}^i, a_{t+2}^i),h_t^j=(a_t^j, s_{t+1}^j, a_{t+1}^j)$.

To see that $T$ is full rank, first permute the rows to be grouped by $a_{t+1}^i$, and permute the columns to be grouped by $a_{t+1}^j$. Then $T$ becomes a block diagonal matrix, with each diagonal block composed of $p( s_{t+2}^i | s_{t+1}^j, a_{t+1})p( a_{t+2}^i | a_t^j)$, where $a_{t+1} = a_{t+1}^i = a_{t+1}^j$. This diagonal block is actually the kronecker product of $T_{a_{t+1}}$ and $\Pi$. Since $T_{a_{t+1}}$ and $\Pi$ are both full rank, their kroncker product is also full rank. Since all the diagonal blocks are full rank, that means $T$ is full rank.

Finally, consider the singular values of $T$. Since $T$ is a block diagonal matrix, the singular values are exactly the singular values of all the diagonal blocks. Each block is kronecker product $T_a \otimes \Pi$, thus the singular values of each block is made up of the product of singular values of $T_a$ and $\Pi$. Therefore $\sigma_1(T) = \max_a \sigma_1(T_a) \sigma_1(\Pi)$ and $\sigma_{\min}(T) = \min_a \sigma_{\min}(T_a) \sigma_{\min}(\Pi)$. Thus
\begin{align}
	\Vert T \Vert_2 &= \max_a \Vert T_a \Vert_2 \Vert \Pi \Vert_2 \\
	&\leq \sqrt{|S|} \\
	\Vert T^{-1} \Vert_2 &= \frac{1}{\sigma_{\min}(T)} \\
	&= \frac{1}{\min_a \sigma_{\min}(T_a) \sigma_{\min}(\Pi)} \\
	&= \frac{\Vert \Pi^{-1} \Vert_2}{\underline{\sigma}_{a}(T_a)} \\
	&\leq \frac{2(1+c|A|)}{\underline{\sigma}_{a}(T_a)}
\end{align}

\subsection{Lemma \ref{lem:app:mom} and Proof}

\begin{applemma}
\label{lem:app:mom}
Given an HMM such that the marginal probability of $h_2$ has positive probability everywhere, the transition matrix is full rank, and the observation matrix is full column rank, then by gathering $N$ samples of $(x_1, x_2, x_3)$, the estimates $\widehat{T},\widehat{O},\widehat{w}$ can be computed such that
\begin{eqnarray}
||\widehat{T} - T||_2 &\leq   18 |A| |S|^{4} (\underline{\sigma}_{a}(R_a) \underline{\sigma}_{a}(Z_{a}))^{-4} \epsilon_1 \\
||\widehat{O} - O||_2 &\leq |A| |S|^{0.5} \epsilon_1\\
||\widehat{O} - O||_{\max} &\leq \epsilon_1\\
||\widehat{w} - w||_2 &\leq 14|A|^2|S|^{2.5} (\underline{\sigma}_{a}(R_a) \underline{\sigma}_{a}(Z_{a}))^{-4} \epsilon_1
\end{eqnarray}
where $||\cdot||_2$ is the spectral norm for matrix arguments, and the Euclidean norm for vector arguments, and $w$ is the marginal probability of $h_2$, with probability $1-\delta$, as long as
\begin{eqnarray}
O \left( \frac{|A|^2 |Z| |R| (1+ \sqrt{\log(1/\delta)})^2}{(C_{d,d,d}(\delta))^2 \cdot \epsilon^2_1} \log \left( \frac{1}{\delta} \right) \right) &\leq N \label{eqn:app:lem2}
\end{eqnarray}
\end{applemma}

\emph{Proof:}

The proof will first use prior method of moments work to derive bounds on the estimation error on $O$ and $OT$ of the induced HMM. Then using those estimates, we will show how to compute an estimate of $T$ and bound its error. Then, we will show how to estimate $w$ and its error. Finally we will give a sufficient lower bound on $N$ in order to achieve an estimation error of $\epsilon_1$ on the estimated matrices.

\subsubsection{Method of Moments}

Theorem 3.1 from \cite{anandkumar2012method} states that given $\epsilon_3$, if the following is satisfied (i.e. there are at least this many samples of $(x_1,x_2,x_3)$)
\begin{align}
	\frac{1+ \sqrt{\log(1/\delta)}}{\sqrt{N}} &\leq C \cdot C_{1,2,3}(\delta) \cdot \epsilon_3 \\
	\iff \frac{1+ \sqrt{\log(1/\delta)}}{C \cdot C_{1,2,3}(\delta) \cdot \epsilon_3} &\leq \sqrt{N} \\
	\iff \frac{(1+ \sqrt{\log(1/\delta)})^2}{C^2 (C_{1,2,3}(\delta))^2 \cdot \epsilon^2_3} &\leq N \label{eq:app:condn}
\end{align}
and also given $\epsilon_2$, if the following is also satisfied
\begin{align}
	\frac{(1+ \sqrt{\log(1/\delta)})^2}{C^2 (C_{3,1,2}(\delta))^2 \cdot \epsilon^2_2} &\leq N \label{eqn:app:mom}
\end{align}
where
\begin{align}
	C_{1,2,3}(\delta) &= \min \left( \frac{\min_{i\neq j}||M_3(\vec{e}_i - \vec{e}_j)||_2 \cdot \sigma_k(P_{1,2})^2}{||P_{1,2,3}||_2 \cdot k^5 \cdot \kappa(M_1)^4} \cdot \frac{\delta}{\log (k / \delta)}, \frac{\sigma_k(P_{1,3})}{1} \right) \\
	C_{1,3,2}(\delta) &= \min \left( \frac{\min_{i\neq j}||M_2(\vec{e}_i - \vec{e}_j)||_2 \cdot \sigma_k(P_{1,3})^2}{||P_{1,3,2}||_2 \cdot k^5 \cdot \kappa(M_1)^4} \cdot \frac{\delta}{\log (k / \delta)} , \frac{\sigma_k(P_{1,2})}{1} \right) \\
	C_{d,d,d}(\delta) &= \min(C_{1,2,3(\delta)},C_{1,3,2}(\delta))
\end{align}
then
\begin{align}
	||col(\tilde{O}) - col(O)||_2 &\leq \epsilon_2 \\
	||col(\tilde{OT}) - col(OT)||_2 &\leq \epsilon_3
\end{align}
with probability $1-\delta$. In other words, the estimated parameters are close in the Euclidean norm. Note that the columns of $\tilde{O}$ and $\tilde{OT}$ may be permuted from $O$ and $OT$, however $\tilde{O}$ and $\tilde{OT}$ are permuted in the same unknown way to match. No knowledge of the permutation is necessary.

Then by matrix norms
\begin{align}
	\Vert \tilde{O} - O \Vert_2 &\leq \Vert \tilde{O} - O \Vert_F \\
	&\leq |A| \sqrt{|S|} \epsilon_2 \\
	\Vert \tilde{O} - O \Vert_{\max} &\leq \max_i ||col_i(\tilde{O}) - col_i(O)||_{\infty} \\
	&\leq \max_i ||col_i(\tilde{O}) - col_i(O)||_2 \\
	&\leq \epsilon_2 \\
	\Vert \tilde{OT} - OT \Vert_2 &\leq |A| \sqrt{|S|} \epsilon_3
\end{align}
using the fact that from the Euclidean distances of the columns we can conclude that the Frobenius norm is $\sqrt{\sum_{ncols} \epsilon_2^2} = \epsilon_2 \sqrt{|A|^2 |S|}$.

\subsubsection{Constructing Transition Matrix Estimate}
\label{sec:app:transition}

The pseudoinverse of $O$ can be used to compute the HMM transition matrix $T$ since $O$ is full column rank
\begin{align}
	O^+ O T &= (O^T O)^{-1}O^T O T \\
	&= T
\end{align}
Then following the same procedure, the HMM transition matrix estimate can be computed
\begin{align}
	\widehat{T} &= \widehat{O}^+ \widehat{OT} \\
	&= (\widehat{O}^T \widehat{O})^{-1} \widehat{O}^T \widehat{OT}
\end{align}
Next, to compute the error
\begin{align}
	||\widehat{T} - T||_2 &= ||(\widehat{O}^T \widehat{O})^{-1} \widehat{O}^T \widehat{OT} - (O^T O)^{-1}O^T O T||_2
\end{align}
first some intermediate quantities will be computed. To begin, because we will be computing inverses, we need to bound the accuracy $\epsilon_2$ in order to use the perturbed inverse lemma; therefore we first assume the following condition
\begin{appcond}
	\begin{align}
		\epsilon_2 &\leq \frac{1}{6|A||S|^{1.5} (\underline{\sigma}_{a}(R_a)\underline{\sigma}_{a}(Z_{a}))^{-4}} \label{eqn:app:cond1}
	\end{align}
\end{appcond}
Then
\begin{align}
	&||\widehat{O}^T \widehat{O} - O^T O||_2 \\
	&\leq ||\widehat{O}^T \widehat{O} - \widehat{O}^T O||_2 + ||\widehat{O}^T O - O^T O||_2 \\
	&\leq ||\widehat{O}^T||_2 || \widehat{O} - O||_2 + ||O||_2 ||\widehat{O}^T  - O^T||_2 \\
	&\leq (||\widehat{O}^T||_2  + ||O||_2 ) |A|\sqrt{|S|}\epsilon_2 \\
	&\leq (2||O||_2 + |A|\sqrt{|S|}\epsilon_2 ) |A|\sqrt{|S|}\epsilon_2 \\
	&\leq (2|S| + 1) |A|\sqrt{|S|}\epsilon_2 \\
	&\leq (3|S|) |A|\sqrt{|S|}\epsilon_2 \\
	&\leq 3|A| |S|^{1.5}\epsilon_2
\end{align}
where we substitute in our previous bounds on the estimation error, matrix norms from lemma \ref{lem:app:hmm}, and also use the reverse triangle inequality on $(||\widehat{O}^T||_2 - ||O||_2 + ||O||_2)$. Then by the perturbed inverse lemma
\begin{align}
	||(\widehat{O}^T \widehat{O})^{-1} - (O^T O)^{-1}||_2 &\leq \frac{3|A| |S|^{1.5}\epsilon_2 ||(O^T O)^{-1}||_2^2}{1 - 3|A| |S|^{1.5}\epsilon_2 ||(O^T O)^{-1}||_2} \\
	&\leq \frac{3|A| |S|^{1.5}\epsilon_2 ||(O^T O)^{-1}||_2^2}{\frac{1}{2}} \\
	&\leq 6|A| |S|^{1.5} ||(O^T O)^{-1}||_2^2 \epsilon_2 \\
	&\leq 6|A| |S|^{1.5} (\underline{\sigma}_{a}(R_a)\underline{\sigma}_{a}(Z_{a}))^{-4} \epsilon_2
\end{align}
where the second inequality follows from substituting in the bound on $\sigma_{\min}(O)$ from lemma \ref{lem:app:hmm} combined with the above condition. Now the error in estimating the HMM transition matrix will be computed
\begin{align}
	& ||\widehat{T} - T||_2 \\
	&= ||(\widehat{O}^T \widehat{O})^{-1} \widehat{O}^T \widehat{OT} - (O^T O)^{-1}O^T (OT)||_2 \\
	&\leq ||(\widehat{O}^T \widehat{O})^{-1} \widehat{O}^T \widehat{OT} - (O^T O)^{-1} \widehat{O}^T \widehat{OT}||_2 + ||(O^T O)^{-1} \widehat{O}^T \widehat{OT} - (O^T O)^{-1}O^T (OT)||_2 \\
	&\leq ||(\widehat{O}^T \widehat{O})^{-1} - (O^T O)^{-1}||_2 || \widehat{O}^T \widehat{OT}||_2 + ||\widehat{O}^T \widehat{OT} - O^T (OT)||_2 ||(O^T O)^{-1}||_2 \\
	&\leq ||(\widehat{O}^T \widehat{O})^{-1} - (O^T O)^{-1}||_2 || \widehat{O}^T ||_2 || \widehat{OT}||_2 \\
	&+ ||\widehat{O}^T  - O^T||_2 ||\widehat{OT}||_2 ||(O^T O)^{-1}||_2 \\
	&+ ||O||_2 || \widehat{OT} - OT||_2 ||(O^T O)^{-1}||_2 \\
	&\leq 6|A| |S|^{1.5} ||(O^T O)^{-1}||_2^2 \epsilon_2 |S|^{2.5} + |A||S|^{0.5}\epsilon_2 |S|^{1.5}||(O^T O)^{-1}||_2 + |S||A||S|^{0.5}\epsilon_3 ||(O^T O)^{-1}||_2
\end{align}
where we substitute in values from lemma \ref{lem:app:hmm} and the estimation errors calculated above.
The next simplification is by letting $\epsilon_2 = \epsilon_3 = \epsilon_1$, then
\begin{align}
	& ||\widehat{T} - T||_2 \\
	&\leq 6|A| |S|^{1.5} ||(O^T O)^{-1}||_2^2 \epsilon_1 |S|^{2.5} + |A||S|^{0.5}\epsilon_1 |S|^{1.5}||(O^T O)^{-1}||_2 + |S||A||S|^{0.5}\epsilon_1 ||(O^T O)^{-1}||_2 \\
	&\leq 6|A| |S|^{4} ||(O^T O)^{-1}||_2^2 \epsilon_1 + |A||S|^{2}||(O^T O)^{-1}||_2 \epsilon_1 + |A||S|^{1.5} ||(O^T O)^{-1}||_2 \epsilon_1\\
	&\leq 6(|A| |S|^{4} + |A||S|^{2} + |A||S|^{1.5}) ||(O^T O)^{-1}||_2^2 \epsilon_1\\
	&\leq 18 |A| |S|^{4} ||(O^T O)^{-1}||_2^2 \epsilon_1\\
	&\leq 18 |A| |S|^{4} (\underline{\sigma}_{a}(R_a)\underline{\sigma}_{a}(Z_{a}))^{-4} \epsilon_1
\end{align}

\subsubsection{Estimating $w$}

Let $X_2$ be random variable for $x_2$ i.e. the HMM observation of the second step. Let $p$ be the  distribution of $x_2$. Let $\widehat{p}$ be the empirical distribution made from the counts of the samples from $x_2$ obtained from the exploring episodes. Then from \cite{weissman2003inequalities} we can bound the deviation
\begin{align}
	\delta_p = P(||\widehat{p} - p||_1 \geq \epsilon_1) &\leq (2^{|A|^2|R||Z|}-2)e^{-\frac{N\epsilon^2}{2}} \\
	\impliedby N \geq O \left( \log \left( \frac{1}{\delta_p} \right)\frac{|A|^2|R||Z|}{\epsilon_1^2} \right)  \label{eq:app:picond}
\end{align}
and get a sufficient condition on how many samples we need. $w=p(h_2)$ can be computed by 
\begin{align}
	p &= Ow \\
	\implies w &= (O)^{+}p \\
	&= ((O)^T(O))^{-1}(O)^T p
\end{align}
and thus estimated using $\widehat{O}$ and $\widehat{p}$. Next, the error of the estimate will be bounded. Let $||\widehat{p} - p||_2 = \epsilon_1$. Note that $||(\widehat{O}^T \widehat{O})^{-1} - (O^T O)^{-1}||_2$ has been computed in the previous section.
\begin{align}
	& ||\widehat{w} - w||_2 \\
	&= ||(\widehat{O}^T \widehat{O})^{-1}\widehat{O}^T \widehat{p} - (O^T O)^{-1}O^T p||_2 \\
	&\leq ||(\widehat{O}^T \widehat{O})^{-1}\widehat{O}^T \widehat{p} - (O^T O)^{-1}\widehat{O}^T \widehat{p}||_2 + ||(O^T O)^{-1}\widehat{O}^T \widehat{p}- (O^T O)^{-1}O^T p||_2 \\
	&\leq ||\widehat{O}^T||_2 ||\widehat{p}||_2||(\widehat{O}^T \widehat{O})^{-1} - (O^T O)^{-1}||_2 + ||(O^T O)^{-1}||_2||\widehat{O}^T \widehat{p}- O^T p||_2 \\
	&\leq ||\widehat{O}^T||_2 ||(\widehat{O}^T \widehat{O})^{-1} - (O^T O)^{-1}||_2 + ||(O^T O)^{-1}||_2(||\widehat{O}^T \widehat{p}-O^T \widehat{p}||_2 + ||O^T \widehat{p}- O^T p||_2) \\
	&\leq (||O||_2 + ||\widehat{O}^T-O^T||_2) ||(\widehat{O}^T \widehat{O})^{-1} - (O^T O)^{-1}||_2 + ||(O^T O)^{-1}||_2(||\widehat{O}^T-O^T||_2 + ||O||_2 \epsilon_1) \\
	&\leq (|S| + |A| |S|^{0.5}\epsilon_1)(6|A||S|^{1.5} ||(O^T O)^{-1}||_2^2 \epsilon_1) + ||(O^T O)^{-1}||_2 (|A||S|^{0.5}\epsilon_1 + |S| \epsilon_1) \\
	&\leq 14|A|^2|S|^{2.5} ||(O^T O)^{-1}||_2^2 \epsilon_1 \\
	&\leq 14|A|^2|S|^{2.5} (\underline{\sigma}_{a}(R_a)\underline{\sigma}_{a}(Z_{a}))^{-4} \epsilon_1
\end{align}
where the first inequality is from using triangle inequality; the third inequality uses the fact that $||\widehat{p}||_1 \leq 1$ and also another triangle inequality; the fifth inequality substitutes in values computed from previous sections; the sixth inequality comes from assuming that $||(O^T O)^{-1}||_2 \leq ||(O^T O)^{-1}||_2^2$, which is implied by assuming that $\underline{\sigma}_{a}(R_a)^{-1} \geq 1$ and $\underline{\sigma}_{a}(Z_{a})^{-1} \geq 1$. To be precise we should instead use the quantities $\max(\underline{\sigma}_{a}(R_a)^{-1}, 1)$ and $\max(\underline{\sigma}_{a}(z_a)^{-1}, 1)$. In the final sample complexity bound, we will implicitly use the versions with the max operator wherever the quantities $\underline{\sigma}_{a}(R_a)^{-1}$ and $\underline{\sigma}_{a}(Z_a)^{-1}$.

\subsubsection{Combining the Conditions}

Recall eqn \ref{eqn:app:mom}
\begin{align}
	\frac{(1+ \sqrt{\log(1/\delta)})^2}{C^2 (C_{3,1,2}(\delta))^2 \cdot \epsilon^2_2} &\leq N
\end{align}
which needs to be satisfied in order to get estimation error bounds on $\epsilon_2$ (and respectively $\epsilon_3$). Combining these requirements with the one from estimating $w$ (eqn \ref{eq:app:picond}), and letting $\epsilon_2 = \epsilon_3 = \epsilon_1$, we get a single requirement of
\begin{align}
	O \left( \frac{|A|^2|R||Z|(1+ \sqrt{\log(3/\delta)})^2}{C^2 (C_{d,d,d}(\delta/3))^2 \cdot \epsilon^2_1} \log \left( \frac{3}{\delta} \right) \right) &\leq N
\end{align}
where we give each of the three requirements an error probability of $\delta/3$.

\subsection{Lemma \ref{lem:app:labelactions} and Proof}

\begin{applemma}
	\label{lem:app:labelactions}
	Given $\widehat{O}$ with max-norm error $\epsilon_O \leq \frac{1}{3|Z||R|}$, then the columns  which correspond to HMM states of the form $h=(a,s',a')$ can be labeled with their corresponding $a,a'$ using Algorithm \ref{alg:app:labelactions}.
\end{applemma}

\emph{Proof:}

Recall $O(i,j) = p(x_t=i | h_t=j) = \delta(a_t^i, a_t^j)\delta(a_{t+1}^i,a_{t+1}^j)p(z_{t+1}^i | a_t^j, s_{t+1}^j)p(r_{t+1}^i | s_{t+1}^j, a_{t+1}^j)$. Consider a column of $O$. The column corresponds to some HMM state $j=(a_t^j,s_{t+1}^j,a_{t+1}^j)$, but the $(a_t^j,s_{t+1}^j,a_{t+1}^j)$ are unknown and only the index $j$ is known. However the rows, which correspond to HMM observation $i=(a_t^i,z_{t+1}^i,r_{t+1}^i,a_{t+1}^i)$, are known since the observations are fully observed. The only entries in this column that can be nonzero are the rows where the actions match i.e. $(a_t,a_{t+1}) = (a_t^i,a_{t+1}^i) = (a_t^j,a_{t+1}^j)$. There are $|Z||R|$ number of these entries. The nonzero entries form a probability distribution and must sum to one. Therefore, the largest nonzero value in this column must be at least $\frac{1}{|Z||R|}$. Since by assumption $\epsilon_O \leq \frac{1}{3|R||Z|}$, the largest nonzero value in $\widehat{O}$ must be at least $\frac{2}{3|Z||R|}$. Thus there exist at least one entry in this column of $\widehat{O}$ that is at least $\frac{2}{3|Z||R|}$ and where the actions match, so it is possible to encounter this during Algorithm \ref{alg:app:labelactions}. Also, Algorithm \ref{alg:app:labelactions} cannot pick entries from $\widehat{O}$ that are zero in $O$, since those entries would be at most only $\frac{1}{3|Z||R|}$. Thus Algorithm \ref{alg:app:labelactions} will always pick an entry with matching actions, and correctly label the HMM state corresponding to each column with the matching actions.

\subsection{Lemma \ref{lem:app:newquantities} and Proof}

\begin{applemma}
\label{lem:app:newquantities}
Given $\widehat{T},\widehat{O},\widehat{w}$ with max-norm 
errors $\epsilon_T,\epsilon_O,\epsilon_w$ respectively, then 
the following bounds hold on the estimated POMDP model parameters 
with probability at least $1-\delta$: 
\begin{align}
&|\widehat{p}(s_{(a',a'')} | s_{(a,a')}, a') - p(s_{(a',a'')} | s_{(a,a')}, a')| \leq \frac{4|S| \epsilon_T}{\epsilon_a^2} \\
& |\widehat{p}(z | a, s_{(a,a')}) - p(z | a, s_{(a,a')})| \leq 4|Z||R|\epsilon_O \\
& |\widehat{p}(r | s_{(a,a')},a') - p(r | s_{(a,a')},a')|  \leq 4|Z||R|\epsilon_O \\
& |\widehat{b}(s_{(a_0,a_1)}) - b(s_{(a_0,a_1)})| \\
&\quad \leq 4|A|^4|S|(||T^{-1}||_2^2 \epsilon_w + 6 ||T^{-1}||_2^3 \epsilon_T)
\end{align}
where $\epsilon_a = \Theta(1/|A|)$
\end{applemma}

\emph{Proof:}

First we will show the errors for the estimated observation, reward, and transition parameters. Then we will show the errors for the estimated initial belief.

\subsubsection{Observation, Reward, and Transition}

By the normalization lemma for $\sum_r \widehat{O}_{HMM,a,a'}((z,r),s(a,a'))$ over $z$
\begin{align}
& |\widehat{p}(z | a, s(a,a')) - p(z | a, s(a,a'))| \\
&= \Vert normalize \left(\sum_r O_{HMM,a,a'}((z,r),s(a,a')) \right) - normalize \left(\sum_r \widehat{O}_{HMM,a,a'}((z,r),s(a,a')) \right)  \Vert_{\infty} \\
&\leq \frac{2( |Z| + 1) |R|\epsilon_O}{1} \\
&= 4|Z||R|\epsilon_O
\end{align}
By the normalization lemma for $\sum_z \widehat{O}_{HMM,a,a'}((z,r),s(a,a'))$ over $r$
\begin{align}
& |\widehat{p}(r | a, s(a,a')) - p(r | a, s(a,a'))| \\
&= \Vert normalize \left(\sum_z O_{HMM,a,a'}((z,r),s(a,a')) \right) - normalize \left(\sum_z \widehat{O}_{HMM,a,a'}((z,r),s(a,a')) \right)  \Vert_{\infty} \\
&\leq \frac{2( |R| + 1) |Z|\epsilon_O}{1} \\
&= 4|Z||R|\epsilon_O
\end{align}
For the transition estimates, first note
\begin{align}
\sum_{s(a',a'')} T_{HMM,a,a',a''}(s(a',a'') | s(a,a')) &= \sum_{s(a',a'')} p(s(a',a'') | s(a',a), a')p( a'' | a) \\
&= p( a'' | a) \\
&= \epsilon_a
\end{align}
Then by the Normalization Lemma
\begin{align}
&|p(s(a',a'') | s(a',a), a') - \widehat{p}(s(a',a'') | s(a',a), a')| \\
&= \Vert normalize(T_{HMM,a,a',a''}(s(a',a'')_i | s(a,a')_j)) - normalize(\widehat{T}_{HMM,a,a',a''}(s(a',a'')_i | s(a,a')_j))\Vert_{\infty} \\
&\leq \frac{2(|S| + \epsilon_a) \epsilon_T}{\epsilon_a^2} \\
&\leq \frac{4|S| \epsilon_T}{\epsilon_a^2}
\end{align}

\subsubsection{Initial distribution}

First, consider $||T^{-1} - \widehat{T}^{-1}||_2$. In order to apply the perturbed inverse lemma, we start be assuming the following condition holds
\begin{appcond}
	\begin{align}
	\epsilon_T &\leq \frac{1}{2||T^{-1}||_2} \label{eqn:app:cond2}
	\end{align}
\end{appcond}
Then it follows by the perturbed inverse lemma that
\begin{align}
||\widehat{T}^{-1} - T^{-1}||_2 &\leq \frac{\epsilon_T ||T^{-1}||_2^2}{1 - ||T^{-1}||_2 \epsilon_T} \\
&\leq 2\epsilon_T ||T^{-1}||_2^2 \\
\end{align}
Then (note $||\cdot||_2$ is Euclidean when the inside is a vector, and the operator norm when the inside is a matrix)
\begin{align}
& ||T^{-1}T^{-1}w - \widehat{T}^{-1}\widehat{T}^{-1}\widehat{w}||_2 \\
&\leq ||T^{-1}T^{-1}w - T^{-1}T^{-1}\widehat{w}||_2 + ||T^{-1}T^{-1}\widehat{w} - \widehat{T}^{-1}\widehat{T}^{-1}\widehat{w}||_2 \\
&\leq ||T^{-1}T^{-1}||_2||w - \widehat{w}||_2 + ||T^{-1}T^{-1} - \widehat{T}^{-1}\widehat{T}^{-1}||_2 ||\widehat{w}||_2 \\
&\leq ||T^{-1}||_2^2 \epsilon_w + ||T^{-1}T^{-1} - \widehat{T}^{-1}\widehat{T}^{-1}||_2 \cdot 1
\end{align}
then that inner term can be simplified
\begin{align}
& ||T^{-1}T^{-1} - \widehat{T}^{-1}\widehat{T}^{-1}||_2 \\
&\leq ||T^{-1}T^{-1} -T^{-1}\widehat{T}^{-1}||_2 + ||T^{-1}\widehat{T}^{-1} - \widehat{T}^{-1}\widehat{T}^{-1}||_2 \\
&\leq ||T^{-1}||_2 ||T^{-1} - \widehat{T}^{-1}||_2 + ||T^{-1} - \widehat{T}^{-1}||_2 ||\widehat{T}^{-1}||_2 \\
&= (||T^{-1}||_2 + ||\widehat{T}^{-1}||_2)||T^{-1} - \widehat{T}^{-1}||_2 \\
&\leq (2||T^{-1}||_2 + ||T^{-1} - \widehat{T}^{-1}||_2)||T^{-1} - \widehat{T}^{-1}||_2 \\
&\leq (2||T^{-1}||_2 + 2\epsilon_T ||T^{-1}||_2^2)2\epsilon_T ||T^{-1}||_2^2 \\
&\leq (2||T^{-1}||_2 + ||T^{-1}||_2)2\epsilon_T ||T^{-1}||_2^2 \\
&\leq 6 ||T^{-1}||_2 \epsilon_T ||T^{-1}||_2^2 \\
&\leq 6 ||T^{-1}||_2^3 \epsilon_T
\end{align}
Thus plugging that back that inner term results in
\begin{align}
||T^{-1}T^{-1}w - \widehat{T}^{-1}\widehat{T}^{-1}\widehat{w}||_2 &\leq ||T^{-1}||_2^2 \epsilon_w + 6 ||T^{-1}||_2^3 \epsilon_T \\
\implies ||T^{-1}T^{-1}w - \widehat{T}^{-1}\widehat{T}^{-1}\widehat{w}||_{\infty} &\leq ||T^{-1}||_2^2 \epsilon_w + 6 ||T^{-1}||_2^3 \epsilon_T \\
\implies |p(a_0, s(a_0,a_1),a_1) - \widehat{p}(a_0, s(a_0,a_1),a_1)| &\leq ||T^{-1}||_2^2 \epsilon_w + 6 ||T^{-1}||_2^3 \epsilon_T
\end{align}
so now the final step is to extract out only the entries corresponding to $(a_0,a_1)$ and normalize (the normalization will be over $s(a_0,a_1)$). Assuming that the following condition holds
\begin{appcond}
	\begin{align}
	(||T^{-1}||_2^2 \epsilon_w + 6 ||T^{-1}||_2^3 \epsilon_T) \leq \frac{1}{2|A|^2|S|} \label{eqn:app:cond3}
	\end{align}
\end{appcond}
then by the Normalization Lemma
\begin{align}
| \widehat{p}(s(a_0,a_1)) - p(s(a_0,a_1))| &= \left| \frac{\widehat{p}(a_0, s(a_0,a_1),a_1)}{\widehat{p}(a_0,a_1)} - \frac{p(a_0, s(a_0,a_1),a_1)}{p(a_0,a_1)} \right| \\
&\leq 2|A|^4(|S| + \frac{1}{|A|^2})(||T^{-1}||_2^2 \epsilon_w + 6 ||T^{-1}||_2^3 \epsilon_T) \\
&\leq 4|A|^4|S|(||T^{-1}||_2^2 \epsilon_w + 6 ||T^{-1}||_2^3 \epsilon_T)
\end{align}
noting that $\sum_{s(a_0,a_1)} p(a_0, s(a_0,a_1),a_1) = p(a_0,a_1)=p(a_0)p(a_1) = \frac{1}{|A|^2}$ (since the first two actions are uniformly random).

\subsection{Lemma \ref{lem:app:betaequiv} and Proof}

\begin{applemma}
	\label{lem:app:betaequiv}
	Given the permutation of the states $s_{(a,a'),j} = s_{\phi((a,a'),j)}$,
	$\beta$-vectors and $\alpha$-vectors over the same policy $\pi_t$ are equivalent i.e.
	\begin{eqnarray}
		\beta_t^{\pi_t}(s_{(a,a'),j}) &= \alpha^{\pi_t}_t(s_{\phi((a,a'),j)})
	\end{eqnarray}
\end{applemma}

\emph{Proof:}

Note that we are using the notation $a_-$ as just another variable for actions, just like the notation $a$. First the base case
\begin{align}
\beta^a_1(s(a_-,a)_j) &= \sum_r p(r|s(a_-,a)_j,a) \cdot r \\
&= \sum_r p(r|s_{\phi(a_-,a,j)},a) \cdot r \\
&= \alpha^a_1(s_{\phi(a_-,a,j)})
\end{align}
Next is the induction step. The induction hypothesis is that this equivalence holds for all $\beta$-vectors with $t$-step policies i.e. $\beta_{t}^{\pi_t}(s(a_-, a)_i) = \alpha_{t}^{\pi_t}(s_{\phi(a_-,a,j)})$ where $a$ is the root action of $\pi_t$. Also note that $f_t(r,z)$ is the rest of conditional policy $\pi_t$ after executing $a$ and seeing $(r,z)$. Then
\begin{align}
& \beta_{t+1}^{a,f_t}(s(a_-, a)_j) \\
&= \sum_{r,z,s(a,f_t(r,z))_k}p(r | s(a_-, a)_j, a)p(s(a,f_t(r,z))_k | s(a_-, a)_j, a)p(z | s(a,f_t(r,z))_k, a)(r + \gamma\beta_{f_t(r,z)}(s(a,f_t(r,z))_k)) \\
&= \sum_{r,z,s(a,f_t(r,z))_k}p(r | s_{\phi(a_-, a,j)}, a)p(s_{\phi(a,f_t(r,z),k)} | s_{\phi(a_-, a,j)}, a)p(z | s_{\phi(a,f_t(r,z),k)}, a)(r + \gamma\alpha_{f_t(r,z)}(s_{\phi(a,f_t(r,z),k)}))
\end{align}
where we use the induction hypothesis.
In order to simplify further, consider the partial term 
\begin{align}
g(r,z) &= \sum_{s(a,f_t(r,z))_k}p(r | s_{\phi(a_-, a,j)}, a)p(s_{\phi(a,f_t(r,z),k)} | s_{\phi(a_-, a,j)}, a)p(z | s_{\phi(a,f_t(r,z),k)}, a)(r + \gamma\alpha_{f_t(r,z)}(s_{\phi(a,f_t(r,z),k)}))
\end{align}
Since this is a sum, the order of summation does not matter. In particular, the summation can be done in the order of the original state $s_i$ where $i=\phi(a,f_t(r,z),k)$.
\begin{align}
g(r,z) &= \sum_{s_i}p(r | s_{\phi(a_-, a,j)}, a)p(s_i | s_{\phi(a_-, a,j)}, a)p(z | s_i, a)(r + \gamma\alpha_{f_t(r,z)}(s_i))
\end{align}
Then
\begin{align}
\beta_{t+1}^{a,f_t}(s(a_-, a)_j) &= \sum_{r,z}g(r,z) \\
&= \sum_{r,z,s_i}p(r | s_{\phi(a_-, a,j)}, a)p(s_i | s_{\phi(a_-, a,j)}, a)p(z | s_i, a)(r + \gamma\alpha_{f_t(r,z)}(s_i)) \\
&= \alpha_{t+1}^{a,f^t}(s_{\phi(a_-, a,j)})
\end{align}

\subsection{Lemma \ref{lem:app:alphaerror} and Proof}

\begin{applemma}
	Suppose we have approximate POMDP parameters with errors
	\begin{eqnarray}
		|\widehat{p}(s' | s, a) - p(s' | s, a)| &\leq \epsilon_T \\
		|\widehat{p}(z | a, s') - p(z | a, s')| &\leq \epsilon_Z \\
		|\widehat{p}(r | s,a) - p(r | s,a')| &\leq \epsilon_R
	\end{eqnarray}
	then for any $t$-step conditional policy  $\pi_t$
	\begin{eqnarray}
		|\alpha^{\pi_t}_{t}(s) - \widehat{\alpha}^{\pi_t}_{t}(s)| &\leq t^2 R_{\max} (|R|\epsilon_R +  |S|\epsilon_T + |Z|\epsilon_Z ).
	\end{eqnarray}
	\label{lem:app:alphaerror}
\end{applemma}

\emph{Proof:}

First the base case for 1-step policies
\begin{align}
|\alpha_1^a(s) - \widehat{\alpha}_1^a(s)| &= \left| \sum_r p(r|s,a) \cdot r - \sum_r \widehat{p}(r|s,a) \cdot r \right| \\
&\leq \sum_r |(p(r|s,a) - \widehat{p}(r|s,a)) \cdot r| \\
&\leq ||p(r|s,a) - \widehat{p}(r|s,a)||_{\infty} \sum_r |r| \\
&\leq |R|R_{\max}\epsilon_R \\
&\leq  R_{\max} (|R|\epsilon_R +  |S|\epsilon_T + |Z|\epsilon_Z )
\end{align}
Next the induction step, where $V_{\max}(t)$ is the upper bound on the value for $t$ steps, and where the induction hypothesis is that $|\alpha_{f_t(r,z)(s')} - \widehat{\alpha}_{f_t(r,z)(s')}| \leq t^2 R_{\max} (|R|\epsilon_R +  |S|\epsilon_T + |Z|\epsilon_Z )$ (also $\gamma=1$)
\begin{align}
& |\alpha_{t+1}^{a,f_t}(s) - \widehat{\alpha}_{t+1}^{a,f_t}(s)| \\
&= \left| \sum_{r,z,s'} p(r | s, a) p(s' | s, a) p(z | s', a) (r + \gamma \alpha_{f_t(r,z)}(s')) - \sum_{r,z,s'} \widehat{p}(r | s, a) \widehat{p}(s' | s, a) \widehat{p}(z | s', a)(r + \gamma \widehat{\alpha}_{f_t(r,z)}(s')) \right| \\
&\leq \sum_{r,z,s'} |p(r | s, a)- \widehat{p}(r | s, a)| p(s' | s, a) p(z | s', a) (r + \gamma \alpha_{f_t(r,z)}(s')) \\
&+ \sum_{r,z,s'} \widehat{p}(r | s, a) |p(s' | s, a)-\widehat{p}(s' | s, a)| p(z | s', a) (r + \gamma \alpha_{f_t(r,z)}(s')) \\
&+ \sum_{r,z,s'} \widehat{p}(r | s, a) \widehat{p}(s' | s, a) |p(z | s', a)-\widehat{p}(z | s', a)| (r + \gamma \alpha_{f_t(r,z)}(s')) \\
&+ \sum_{r,z,s'} \widehat{p}(r | s, a) \widehat{p}(s' | s, a) \widehat{p}(z | s', a) |(r + \gamma \alpha_{f_t(r,z)}(s')) - (r + \gamma \widehat{\alpha}_{f_t(r,z)}(s'))| \\
&\leq V_{\max}(t+1) \sum_{r} |p(r | s, a)- \widehat{p}(r | s, a)| \sum_{s'} p(s' | s, a)\sum_{z}p(z|s',a) \\
&+ V_{\max}(t+1) \sum_{s'} |p(s' | s, a)-\widehat{p}(s' | s, a)| \sum_z p(z | s', a) \sum_r \widehat{p}(r | s, a) \\
&+ V_{\max}(t+1) \sum_{r} \widehat{p}(r | s, a) \sum_{s'} \widehat{p}(s' | s, a) \sum_{z} |p(z | s', a)-\widehat{p}(z | s', a)|  \\
&+ \gamma \sum_{r} \widehat{p}(r | s, a) \sum_{s'} \widehat{p}(s' | s, a) \sum_{z} \widehat{p}(z | s', a) |\alpha_{f_t(r,z)}(s')) - \widehat{\alpha}_{f_t(r,z)}(s')| \\
&\leq V_{\max}(t+1) (|R|\epsilon_R +  |S|\epsilon_T + |Z|\epsilon_Z ) + \gamma(t^2 R_{\max} (|R|\epsilon_R +  |S|\epsilon_T + |Z|\epsilon_Z )) \\
&\leq (t+1) R_{\max} (|R|\epsilon_R +  |S|\epsilon_T + |Z|\epsilon_Z ) + t^2 R_{\max} (|R|\epsilon_R +  |S|\epsilon_T + |Z|\epsilon_Z ) \\
&\leq (t^2 + t +1) R_{\max} (|R|\epsilon_R +  |S|\epsilon_T + |Z|\epsilon_Z ) \\
&\leq (t+1)^2 R_{\max} (|R|\epsilon_R +  |S|\epsilon_T + |Z|\epsilon_Z )
\end{align}

\subsection{Lemma \ref{lem:app:findpolicy} and Proof}

\begin{applemma}
	\label{lem:app:findpolicy}
	Algorithm \ref{app:alg:findpolicy} finds the policy $\widehat{\pi}$ which maximizes $V^{\widehat{\pi}}(\widehat{b}(s_1))$ for a POMDP with parameters
	%\begin{eqnarray*}
	$\widehat{b}(s_1), \widehat{p}(z | a, s'), \widehat{p}(r | s,a),$ and $\widehat{p}(s' | s, a)$.
	%\end{eqnarray*}
\end{applemma}

\emph{Proof:}

The outer loop builds up the set $\Gamma_t^{a_-,a}$ of beta vectors that take as input $s(a_-,a)$ with $a$ as the root action. This is true in the base case for $\Gamma_1^{a_-,a}$. For the induction step, fix $(a_-,a$. Then $f_t(r,z)$ is taken from all possible mappings from an observation pair to $\beta_{t-1}(s(a,a')) \in \Gamma_{t-1}^{a,a'}$ and all possible next actions $a'$. Thus all possible $\beta_t(s(a_-,a))$ are computed.

The final step is an argmax. Since $\widehat{p}(s_1(a_0,a_1))$ is just a permutation of $\widehat{p}(s_1)$, using it will not change the dot product $\widehat{p}(s_1(a_0,a_1)) \cdot \beta_H(s(a_0,a_1))$. Thus the argmax is correctly finding the policy associated with $\widehat{V}^*(\widehat{p}(s_1))$.

\section{Main Theorem}
\label{sec:app:theorem}

\begin{apptheorem}
	For POMDPs that satisfy the stated assumptions defined in the 
	problem setting, executing \porl\ 
	%Executing Algorithm \ref{alg:app:pac} 
	will achieve an expected episodic reward of $V(b_0) \geq V^*(b_0) - \epsilon$ 
	on all but a number of episodes that is bounded by 
	\begin{eqnarray*}
		O \left( \frac{H^4 V_{\max}^2|A|^{12} |R|^{4}|Z|^{4}|S|^{12}  \left(1+ \sqrt{\log \frac{3}{\delta}}\right)^2}{ C_{d,d,d}\left(\frac{\delta}{3}^2\right) \underline{\sigma}_{a}(T_a)^{6} \underline{\sigma}_{a}(R_a)^8 \underline{\sigma}_{a}(Z_{a})^{8}  \epsilon^2} \log \left( \frac{3}{\delta} \right) \right)
	\end{eqnarray*}
	with probability at least $1-\delta$, where  
	%bad episodes in which $V(b) \leq V^*(b) - \epsilon$ with probability $1-\delta$ where
	\[
	C_{d,d,d}(\delta) = \min(C_{1,2,3}(\delta),C_{1,3,2}(\delta))
	\]
	where
	\begin{eqnarray*}
		C_{1,2,3}(\delta) &=& \min \left( \frac{\min_{i\neq j}||M_3(\vec{e}_i - \vec{e}_j)||_2  \cdot \sigma_k(P_{1,2})^2}{||P_{1,2,3}||_2 \cdot k^5 \cdot \kappa(M_1)^4} \cdot \frac{\delta}{\log (k / \delta)}, \frac{\sigma_k(P_{1,3})}{1} \right) \\
		C_{1,3,2}(\delta) &=& \min \left( \frac{\min_{i\neq j}||M_2(\vec{e}_i - \vec{e}_j)||_2  \cdot \sigma_k(P_{1,3})^2}{||P_{1,3,2}||_2 \cdot k^5 \cdot \kappa(M_1)^4} \cdot \frac{\delta}{\log (k / \delta)} , \frac{\sigma_k(P_{1,2})}{1} \right) \\
	\end{eqnarray*}
\end{apptheorem}

\emph{Proof:}

Let the initial beliefs $b,\widehat{b}$ and error $\Vert b - \widehat{b} \Vert_{\infty} \leq \epsilon_b$, and the bound over $\alpha$-vectors of any policy $\pi$, $\Vert \alpha^{\pi} - \widehat{\alpha}^{\pi} \Vert_{\infty} \leq \epsilon_{\alpha}$ be given. Let $\widehat{\pi}$ be the policy returned by Algorithm \ref{app:alg:findpolicy} i.e the optimal policy for $\widehat{b}$ and $\widehat{\alpha}$. Let $\pi^*$ be the optimal policy for $b$ and $\alpha$. Then
\begin{align}
\widehat{V}^{\widehat{\pi}}(\widehat{b}) &= \widehat{b} \cdot \widehat{\alpha}^{\widehat{\pi}} \\
&\geq \widehat{b} \cdot \widehat{\alpha}^{\pi^*} \\
&\geq \widehat{b} \cdot \alpha^{\pi^*} - |\widehat{b} \cdot \alpha^{\pi^*} -  \widehat{b} \cdot \widehat{\alpha}^{\pi^*}| \\
&\geq \widehat{b} \cdot \alpha^{\pi^*} - \epsilon_{\alpha} \\
&\geq b \cdot \alpha^{\pi^*} - |b \cdot \alpha^{\pi^*} - \widehat{b} \cdot \alpha^{\pi^*}| - \epsilon_{\alpha} \\
&\geq b \cdot \alpha^{\pi^*} - \epsilon_{b} V_{\max} - \epsilon_{\alpha} \\
&= V^*(b) - \epsilon_{b} V_{\max} - \epsilon_{\alpha}
\end{align}
where the first inequality is because $\widehat{\pi}$ is the optimal policy for $\widehat{b}$; the second inequality comes from triangle inequality; the third inequality uses Holder's inequality and the fact that $||\widehat{b}||_{\infty} \leq 1$; the fourth inequality uses triangle inequality again. Next
\begin{align}
V^{\widehat{\pi}}(b) &= b \cdot \alpha^{\widehat{\pi}} \\
&\geq \widehat{b} \cdot \alpha^{\widehat{\pi}} - |\widehat{b} \cdot \alpha^{\widehat{\pi}} - b \cdot \alpha^{\widehat{\pi}}| \\
&\geq \widehat{b} \cdot \alpha^{\widehat{\pi}} - \epsilon_{b}V_{\max} \\
&\geq \widehat{b} \cdot \widehat{\alpha}^{\widehat{\pi}} - |\widehat{b} \cdot \widehat{\alpha}^{\widehat{\pi}} - \widehat{b} \cdot \alpha^{\widehat{\pi}}| - \epsilon_{b}V_{\max} \\
&\geq \widehat{b} \cdot \widehat{\alpha}^{\widehat{\pi}} - \epsilon_{\alpha} - \epsilon_{b}V_{\max}
\end{align}
Putting those two together results in
\begin{align}
V^{\widehat{\pi}}(b) &\geq V^*(b) - 2\epsilon_{b} V_{\max} - 2\epsilon_{\alpha}
\end{align}
Plugging in $\epsilon_{b}$ and $\epsilon_{\alpha}$ from lemma \ref{lem:app:newquantities} and lemma \ref{lem:app:alphaerror} gets us
\begin{align}
V^{\widehat{\pi}}(b) &\geq V^*(b) - 2(4|A|^4|S|(||T^{-1}||_2^2 \epsilon_w + 6 ||T^{-1}||_2^3 \epsilon_T)) V_{\max} - 2(H^2 R_{\max} (|R|\epsilon_R +  |S|\epsilon_T + |Z|\epsilon_Z ))
\end{align}
We know $||T^{-1}||_2 = 2(1+c|A|)(\underline{\sigma}_{a}(T_a))^{-1}$ and $\epsilon_a = \frac{c}{1+c|A|}$ from Lemma \ref{lem:app:hmm} and the exploration policy lemma. Note that $c=O(1/|A|)$ and so $\epsilon_a = O(1/|A|)$.
Now let's carefully substitute in quantities from Lemma \ref{lem:app:mom} and Lemma \ref{lem:app:newquantities}
\begin{align}
& (||T^{-1}||_2^2 \epsilon_w + 6 ||T^{-1}||_2^3 \epsilon_T) \\
&\leq ||T^{-1}||_2^2(14|A|^2|S|^{2.5} (\underline{\sigma}_{a}(R_a)\underline{\sigma}_{a}(Z_{a}))^{-4} \epsilon_1) \\
&+ 6 ||T^{-1}||_2^3 (18 |A| |S|^{4} (\underline{\sigma}_{a}(R_a)\underline{\sigma}_{a}(Z_{a}))^{-4} \epsilon_1) \\
&\leq 122||T^{-1}||_2^3 |A|^2|S|^{4} (\underline{\sigma}_{a}(R_a)\underline{\sigma}_{a}(Z_{a}))^{-4} \epsilon_1 \\
&\leq 122(8(1+c|A|)^3(\underline{\sigma}_{a}(T_a)^{-3}) |A|^2|S|^{4} (\underline{\sigma}_{a}(R_a)\underline{\sigma}_{a}(Z_{a}))^{-4} \epsilon_1 \\
&\leq O \left( |A|^2|S|^{4} \underline{\sigma}_{a}(T_a)^{-3} (\underline{\sigma}_{a}(R_a)\underline{\sigma}_{a}(Z_{a}))^{-4} \epsilon_1 \right) \label{eqn:app:cond3val}
\end{align}
\begin{align}
& (|R|\epsilon_R +  |S|\epsilon_T + |Z|\epsilon_Z ) \\
&\leq (|R|(4|Z||R|\epsilon_O) +  |S|\frac{4|S|\epsilon_T}{\epsilon_a^2} + |Z|(4|Z||R|\epsilon_O) ) \\
&\leq |R|(4|Z||R|\epsilon_1) +  |S|\frac{18 \cdot 4}{\epsilon_a^2}|S| |A| |S|^{4} (\underline{\sigma}_{a}(R_a) \underline{\sigma}_{a}(Z_{a}))^{-4} \epsilon_1 + |Z|4|Z||R|\epsilon_1 \\
&\leq (18 \cdot 4 + 8) \frac{1}{\epsilon_a^2} |A||S|^{6} |R|^{2}|Z|^{2} (\underline{\sigma}_{a}(R_a) \underline{\sigma}_{a}(Z_{a}))^{-4} \epsilon_1  \\
&= O\left( |A|^3|S|^{6} |R|^{2}|Z|^{2} (\underline{\sigma}_{a}(R_a) \underline{\sigma}_{a}(Z_{a}))^{-4} \epsilon_1 \right)
\end{align}
now putting that back into the bound
\begin{align}
V^{\widehat{\pi}}(b) &\geq V^*(b) \\
&- O \left( V_{\max}|A|^{6}|S|^{5} \underline{\sigma}_{a}(T_a)^{-3} (\underline{\sigma}_{a}(R_a)\underline{\sigma}_{a}(Z_{a}))^{-4} \epsilon_1 \right) \\
&- O\left(H^2 R_{\max}|A|^4|S|^{6} |R|^{2}|Z|^{2} (\underline{\sigma}_{a}(R_a) \underline{\sigma}_{a}(Z_{a}))^{-4} \epsilon_1 \right) \\
&- O \left( H^2 V_{\max}|A|^{5} |R|^{2}|Z|^{2}|S|^{6} (\underline{\sigma}_{a}(T_a))^{-3} (\underline{\sigma}_{a}(R_a)\underline{\sigma}_{a}(Z_{a}))^{-4} \epsilon_1 \right)
\end{align}
now if we let that error be equal to $\epsilon$ i.e. $V^{\widehat{\pi}}(b) \geq V^*(b) - \epsilon$, then we can substitute $\epsilon$ for $\epsilon_1$ into eqn \ref{eqn:app:lem2} from Lemma \ref{lem:app:mom} to get the following requirement on $N$
\begin{align}
O \left( \frac{H^4 V_{\max}^2|A|^{12} |R|^{4}|Z|^{4}|S|^{12} (\underline{\sigma}_{a}(T_a))^{-6} (\underline{\sigma}_{a}(R_a)\underline{\sigma}_{a}(Z_{a}))^{-8} (1+ \sqrt{\log(3/\delta)})^2}{ (C_{d,d,d}(\delta/3))^2 \cdot \epsilon^2} \log \left( \frac{3}{\delta} \right) \right) &\leq N \label{eqn:app:final}
\end{align}
where
\begin{align}
C_{d,d,d}(\delta) &= \min(C_{1,2,3(\delta)},C_{1,3,2}(\delta)) \\
C_{1,2,3}(\delta) &= \min \left( \frac{\min_{i\neq j}||M_3(\vec{e}_i - \vec{e}_j)||_2 \cdot \sigma_k(P_{1,2})^2}{||P_{1,2,3}||_2 \cdot k^5 \cdot \kappa(M_1)^4} \cdot \frac{\delta}{\log (k / \delta)}, \frac{\sigma_k(P_{1,3})}{1} \right) \\
C_{1,3,2}(\delta) &= \min \left( \frac{\min_{i\neq j}||M_2(\vec{e}_i - \vec{e}_j)||_2 \cdot \sigma_k(P_{1,3})^2}{||P_{1,3,2}||_2 \cdot k^5 \cdot \kappa(M_1)^4} \cdot \frac{\delta}{\log (k / \delta)} , \frac{\sigma_k(P_{1,2})}{1} \right)
\end{align}
are the quantities from MoM from \cite{anandkumar2012method}.

And that is the final sample complexity bound (eqn \ref{eqn:app:final}). Note that we assume $\underline{\sigma}_{a}(R_a)^{-1} \geq 1$, $\underline{\sigma}_{a}(Z_a)^{-1} \geq 1$ and $\underline{\sigma}_{a}(T_a)^{-1} \geq 1$ or else we can just replace those quantities by $1$ in the bound.

One final thing to do is to give sufficient conditions on $N$ so that the conditions made on all of the $\epsilon_1, \epsilon_2, \dots$ in the lemmas hold. We then note that the final sample complexity bound is a sufficient condition on $N$ (we only consider $\epsilon \leq 1$ to be interesting and don't consider $\epsilon > 1$).

From Lemma \ref{lem:app:labelactions}, the condition is that $\epsilon_O \leq \frac{1}{3|R||Z|}$. This translates into 
\begin{align}
\epsilon_1 \leq \frac{1}{3|R||Z|} \\
\impliedby O \left( \frac{|A|^2 |Z|^2 |R|^2 (1+ \sqrt{\log(3/\delta)})^2}{(C_{d,d,d}(\delta/3))^2} \log \left( \frac{3}{\delta} \right) \right) &\leq N
\end{align}
where the first inequality is due to $\epsilon_1 = \epsilon_O$ from Lemma \ref{lem:app:mom}; the second inequality is from substituting the first inequality into eqn \ref{eqn:app:lem2} from Lemma $\ref{lem:app:mom}$. The second inequality condition is already satisfied by the final bound (eqn \ref{eqn:app:final}). Next up is the condition (eqn \ref{eqn:app:cond1}) made during the proof of Lemma \ref{lem:app:mom} where 
\begin{align}
\epsilon_1 \leq \frac{1}{6|A||S|^{1.5} (\underline{\sigma}_{a}(R_a)\underline{\sigma}_{a}(Z_{a}))^{-4}} \\
\impliedby O \left( \frac{|A|^4 |S|^3 |Z| |R| (\underline{\sigma}_{a}(R_a)\underline{\sigma}_{a}(Z_{a}))^{-8} (1+ \sqrt{\log(3/\delta)})^2}{(C_{d,d,d}(\delta/3))^2} \log \left( \frac{3}{\delta} \right) \right) \leq N
\end{align}
where the second inequality is obtrained from substituting the first inequality into eqn \ref{eqn:app:lem2}. The second inquality is also already satisfied by the final bound. Then the condition made during the proof of Lemma \ref{lem:app:newquantities}  (eqn \ref{eqn:app:cond2}) is
\begin{align}
\epsilon_T \leq \frac{1}{2||T^{-1}||_2} \\
\impliedby \epsilon_1 \leq \frac{1}{36 |A| |S|^{4} (\underline{\sigma}_{a}(R_a) \underline{\sigma}_{a}(Z_{a}))^{-4} (\frac{2(1+c|A|)}{\underline{\sigma}_{a}(T_a)})} \\
\impliedby \epsilon_1 \leq O\left( \frac{1}{|A| |S|^{4} (\underline{\sigma}_{a}(R_a) \underline{\sigma}_{a}(Z_{a}))^{-4} (\underline{\sigma}_{a}(T_a))^{-1}} \right) \\
\impliedby O \left( \frac{|A|^4 |S|^8 |Z| |R| (\underline{\sigma}_{a}(R_a)\underline{\sigma}_{a}(Z_{a}))^{-8} (\underline{\sigma}_{a}(T_a))^{-2}(1+ \sqrt{\log(3/\delta)})^2}{(C_{d,d,d}(\delta/3))^2} \log \left( \frac{3}{\delta} \right) \right) \leq N
\end{align}
where the second inequality comes from substituting the value for $||T^{-1}||_2$ from Lemma \ref{lem:app:hmm}, and then using the relationship between $\epsilon_1$ and $\epsilon_T$ from Lemma \ref{lem:app:mom}. The fourth inequality is from eqn \ref{eqn:app:lem2} and is also satisfied by the final bound. Finally the second condition from the proof of Lemma \ref{lem:app:newquantities} (eqn \ref{eqn:app:cond3}) is
\begin{align}
(||T^{-1}||_2^2 \epsilon_w + 6 ||T^{-1}||_2^3 \epsilon_T) \leq \frac{1}{2|A|^2|S|} \\
\impliedby \epsilon_1 \leq O \left( \frac{1}{|A|^4|S|^{5} \underline{\sigma}_{a}(T_a)^{-3} (\underline{\sigma}_{a}(R_a)\underline{\sigma}_{a}(Z_{a}))^{-4} } \right) \\
\impliedby O \left( \frac{|A|^{10} |S|^{10} |Z| |R| \underline{\sigma}_{a}(T_a)^{-6} (\underline{\sigma}_{a}(R_a)\underline{\sigma}_{a}(Z_{a}))^{-8} (1+ \sqrt{\log(3/\delta)})^2}{(C_{d,d,d}(\delta/3))^2} \log \left( \frac{3}{\delta} \right) \right) \leq N
\end{align}
where the second inequality comes from using eqn \ref{eqn:app:cond3val}. The third inequality is from eqn \ref{eqn:app:lem2} and is also satisfied by the final bound. Thus, the given final sample complexity bound is sufficient.

	}
	
\end{document}